\documentclass[10pt,journal,compsoc]{IEEEtran}
\usepackage{ifpdf}
 \ifpdf
 \else
 \fi

\ifCLASSOPTIONcompsoc
  \usepackage[nocompress]{cite}
\else
  \usepackage{cite}
\fi

\ifCLASSINFOpdf
\else
\fi

\usepackage{amsmath}
\usepackage{algorithmic}
\usepackage{array}
\usepackage{graphicx}
\usepackage{multirow}
\usepackage{booktabs}
\usepackage{caption}
\usepackage{ragged2e}
\usepackage{bm}

\ifCLASSOPTIONcompsoc
  \usepackage[caption=false,font=footnotesize,labelfont=sf,textfont=sf]{subfig}
\else
  \usepackage[caption=false,font=footnotesize]{subfig}
\fi

\usepackage{fixltx2e}
\usepackage{stfloats}

\ifCLASSOPTIONcaptionsoff
  \usepackage[nomarkers]{endfloat}
 \let\MYoriglatexcaption\caption
 \renewcommand{\caption}[2][\relax]{\MYoriglatexcaption[#2]{#2}}
\fi

\usepackage{url}

\usepackage{mathrsfs}
\usepackage{amssymb}

\begin{document}
%
\title{Learning to Infer Unseen Single-/Multi-Attribute-Object Compositions with Graph Networks}
%
%
%
%

\author{Hui~Chen,
        Jingjing~Jiang,
        and~Nanning~Zheng,~\IEEEmembership{Fellow,~IEEE}
\IEEEcompsocitemizethanks{\IEEEcompsocthanksitem H.Chen, J. Jiang and N. Zheng are with the Institute of Artificial Intelligence and Robotics, Xi’an Jiaotong University, Shaanxi, 710049 China (e-mail: chenhui0622@stu.xjtu.edu.cn; jingjingjiang2017@gmail.com; nnzheng@mail.xjtu.edu.cn). N. Zheng is the corresponding author.
}
}

\IEEEtitleabstractindextext{%
\begin{abstract}
\justifying
Inferring the unseen attribute-object composition is critical to make machines learn to decompose and compose complex concepts like people. Most existing methods are limited to the composition recognition of single-attribute-object, and can hardly learn relations between the attributes and objects. In this paper, we propose an attribute-object semantic association graph model to learn the complex relations and enable knowledge transfer between primitives. With nodes representing attributes and objects, the graph can be constructed flexibly, which realizes both single- and multi-attribute-object composition recognition. In order to reduce mis-classifications of similar compositions (e.g., \texttt{scratched screen} and \texttt{broken screen}), driven by the contrastive loss, the anchor image feature is pulled closer to the corresponding label feature and pushed away from other negative label features. Specifically, a novel balance loss is proposed to alleviate the domain bias, where a model prefers to predict seen compositions. In addition, we build a large-scale Multi-Attribute Dataset (MAD) with 116,099 images and 8,030 label categories for inferring unseen multi-attribute-object compositions. Along with MAD, we propose two novel metrics Hard and Soft to give a comprehensive evaluation in the multi-attribute setting. Experiments on MAD and two other single-attribute-object benchmarks (MIT-States and UT-Zappos50K) demonstrate the effectiveness of our approach.
\end{abstract}

}

\maketitle

\IEEEdisplaynontitleabstractindextext

%
\IEEEpeerreviewmaketitle

\IEEEraisesectionheading{\section{Introduction}\label{sec:introduction}}

%
%
%
%
\IEEEPARstart{T}{he} ultimate goal of AI is to make machines learn and think like people. Recently, the remarkable developments \cite{he2016deep,silver2016mastering,silver2017mastering} of deep learning \cite{lecun2015deep} have quickened the progress toward this goal. However, machine intelligence is still far from matching human intelligence. Humans can learn richer semantic representations from less data and generalize in a more flexible way. Usually, only one or a few examples are needed before a human being can learn a concept that can be flexibly applied in other domains (e.g., generalizing known concepts to recognize new examples or composing multiple concepts to understand new concepts). To build a machine to learn, compose and generalize concepts like humans on the basis of the visual features of object entity in the image, this paper handles a challenging problem, which is called unseen attribute-object composition recognition. Given an image, unseen attribute-object composition recognition asks a model to simultaneously recognize the attribute type and object type of the image even though that attribute-object composition is not included in the training set. For example, the training set includes samples of \texttt{sliced tomato} and \texttt{green apple} (seen compositions), while the testing set includes samples of \texttt{sliced apple} (unseen composition). To correctly recognize unseen compositions, the model needs to learn the concepts of attributes and objects from the seen compositions and then compose and generalize the concepts to recognize unseen compositions.

Unseen attribute-object composition recognition is a meaningful problem. From the perspective of cognitive psychology, the attribute is a high-level semantic description for objects, which is closely related to the formation of abstract concepts, particularly the multi-attribute cognition of objects plays an important role in understanding the environment in people’s daily life \cite{solso2005cognitive}. The ability to recognize unseen compositions by inferring and generalizing assists human beings to understand new objects. Therefore, building a machine to recognize unseen attribute-object compositions like people can achieve a deeper understanding of images, allowing for advanced applications, such as image retrieval \cite{johnson2015image}, pedestrian re-identification \cite{lin2019improving}, scene graph generation \cite{xu2017scene}, relation reasoning \cite{santoro2017a}, and object perception \cite{devereux2018integrated,lake2015human-level}. For example, compared with a system that can recognize only objects, a system that understands the concepts of attribute-object compositions can output more accurate retrieval results. In addition, recognizing unseen compositions based on those that are seen requires only data annotations of the seen compositions, relieving the resource consumption of data annotations for all compositions.

The main challenges of unseen attribute-object composition recognition are as follows: 1) The attribute is abstract because we cannot find a proper prototype even for a simple attribute (e.g., \texttt{beautiful}). In addition, the relations between the attributes and objects are complex. The same attribute presents diverse visual appearances in different attribute-object compositions (e.g., \texttt{old dog} and \texttt{old bridge}), making it difficult to learn an intrinsic feature representation of an attribute. 2) Some different compositions (e.g., \texttt{scratched screen} and \texttt{broken screen}) share similar appearances in such a way that they are easily misclassified. In other words, it is difficult to learn discriminative features. 3) There lacks a benchmark dataset describing the multiple attributes of objects. Existing single-attribute annotations \cite{isola2015discovering,yu2014fine-grained} may be one-sided and inaccurate for the co-occurrence patterns of the attributes (e.g., \texttt{old} and \texttt{rusty} for \texttt{car}).

Methods for unseen attribute-object composition recognition have been active throughout the transition from discriminative models to generative models. In the early times, discriminative methods typically learned attribute and object classifiers in the beginning, and then, they recognized unseen compositions by composing these separately-trained classifiers \cite{chen2014inferring,elhoseiny2013write,lu2016visual,misra2017from}. These methods largely push forward the study of unseen attribute-object composition recognition, but separately-trained classifiers ignore the relations between the attributes and objects. Unseen attribute-object composition recognition is a high-level and complex vision problem that relies on a deep understanding of an image, which asks a model to consider the relations between the attributes and objects. Recent studies \cite{nagarajan2018attributes,purushwalkam2019task-driven,li2020symmetry,naeem2021learning} have attempted to process the attribute and object as a whole to explore their inner relations. For example, Nagarajan \textit{et al}. \cite{nagarajan2018attributes} regard an attribute as a linear transformation matrix and an object as an embedding vector, and an attribute-object composition is finally formulated as the object vector multiplied by the attribute matrix. More recently, generative models \cite{kodirov2017semantic,nan2019recognizing,wei2019adversarial,li2020symmetry} have achieved state-of-the-art performance. For example, Nan \textit{et al}. \cite{nan2019recognizing} learn the attribute-object relations with an encoder-decoder based generative model. Wei \textit{et al}. \cite{wei2019adversarial} utilize the positive and (semi-) negative attribute-object composition embeddings to learn more accurate correlations between the attributes and objects. Although these methods have achieved impressive progress, two problems remain to be solved: 1) How to design an effective model to learn the inner relations between the attributes and objects, so as to learn better primitive and composition embeddings, and 2) how to distinguish the attribute-object compositions with similar appearances.

In this work, two kinds of data modalities including visual images and the corresponding linguistic labels, are involved in unseen attribute-object composition recognition. Therefore, we propose a two-pathway model to encode the two kinds of data separately. In the visual pathway, we first extract the initial visual feature vectors of input images via pre-trained convolutional neural networks \cite{he2016deep}. Then, we project the visual features to a unit hyper-sphere in a latent space, obtaining the encoded visual features. In the linguistic pathway, to learn the interdependence between the attributes and objects, we propose an attribute-object semantic association graph network to enable knowledge transfer between nodes. Each node in the graph represents the linguistic feature of an attribute or object entity, iteratively aggregating the information of neighborhood nodes, and the graph network outputs a node embedding matrix in the latent space (each row denotes the feature of a graph node). The motivations for using a graph to encode linguistic information are two-fold: 1) The complex interdependence between the attributes and objects can be modeled via the graph, in which a large connection strength indicates an obvious semantic correlation between two nodes. 2) A graph is flexible and extensible while handling multi-attribute-object composition recognition. In general, the attribute and object are two fundamentally different entities with the attribute being more abstract. Therefore, after the graph network, an attention-based simple composer module is responsible for fusing the attribute and object embeddings instead of adding them together directly, which outputs a holistic composition representation.

To train our model, we first utilize the contrastive loss \cite{he2020momentum,pmlr-v119-chen20j} to learn discriminative features by maximizing the similarity between an anchor image and the correct composition while minimizing the similarities with other incorrect compositions. Besides, we propose a novel balance loss to tackle the challenge of domain bias in generalized unseen attribute-object recognition \cite{purushwalkam2019task-driven}, aiming to raise the model's awareness of all unseen compositions. We evaluate our model on two commonly-used benchmark datasets, i.e., MIT-States \cite{isola2015discovering} and UT-Zappos50K \cite{yu2014fine-grained}. On both datasets, our model outperforms the state-of-the-art baseline methods. Furthermore, we build a large-scale Multi-Attribute Dataset (MAD) to describe the multiple attributes of objects. Along with MAD, we propose two novel metrics Hard and Soft from the perspective of compositionality to give a comprehensive evaluation in the multi-attribute setting. On this highly complex dataset with large variations in attribute and object distributions, our model achieves competitive recognition results.

The main contributions of this paper can be summarized below. 1) We extend the current research scope of unseen attribute-object composition recognition from single-attribute to multi-attribute case which is more challenging and realistic. Moreover, we propose an end-to-end neural network model to handle both single- and multi-attribute-object composition recognition tasks, which shows state-of-the-art performance. 2) We construct a semantic association graph network over the attributes and objects to learn compositional representations. The graph is mapped to a set of interdependent node representations to explore the inner relations between the attributes and objects. 3) We propose a novel balance loss to tackle the domain bias problem in generalized unseen attribute-object recognition, which includes both the seen and unseen compositions during the testing. 4) We built a multi-attribute dataset called MAD. To the best of our knowledge, MAD is the first dataset for inferring unseen multi-attribute-object compositions. Further, we propose two novel metrics Hard and Soft to give a comprehensive performance evaluation in the multi-attribute setting. We hope that MAD can promote further developments of compositional learning and other related fields in the computer vision community.

We will publicly release the code and dataset associated with this paper.

\section{Related Work}
\subsection{Zero-shot Learning} 
Zero-shot Learning (ZSL) aims to recognize unseen concepts (usually referring to objects) after learning only from training seen concepts and additional high-level semantic information (e.g., attributes and text). In the ZSL setting, the training label set and testing label set are disjoint. Early works \cite{lampert2009learning,lampert2014attribute-based} typically extract visual feature in the visual space and project it into a semantic space where the attribute descriptions of unseen objects are known, and the recognition is implemented by searching the object whose description mostly matches the visual feature. For these methods, attribute classifiers are trained separately and the relations between attributes are ignored. In order to mitigate this issue, some works \cite{li2018discriminative,wang2019visual} seek to embed the visual features and the attribute descriptions into a common latent space. However, the above methods lack either the ability to learn the bi-directional mappings between the visual space and the semantic/latent space or a flexible metric to evaluate the similarity between different kinds of features. Huang \textit{et al.} \cite{huang2018generative} leverage generative adversarial networks (GAN) to generate various visual features conditioned on class labels and map each visual feature to its corresponding semantic feature. Apart from learning mappings between the visual space and semantic space, some useful techniques have been introduced in ZSL to improve recognition performance, such as memory mechanism \cite{santoro2017a}, knowledge graph \cite{wang2018zero-shot,kampffmeyer2019rethinking} and low-rank constraint \cite{liu2018zero}. A good survey of ZSL can be seen in \cite{2019Zero}.

\subsection{Graph Convolutional Networks}
Researchers have proposed to tackle object recognition problem via a graph to exploit complex relations between entities \cite{chen2019multi-label,gao2019graph,liu2019independence,vashishth2019composition-based,velickovic2017graph}. Let $\mathcal{G} = (\mathcal{V}, \mathcal{A}, Z)$ denote a graph, where $\mathcal{V}$ denotes the set of $N$ nodes and $\mathcal{A}$ is the adjacency matrix of the graph, which represents connection relations between edges. $Z \in \mathbb{R}^{N \times d}$ is a matrix that encodes node features, and $Z_i$ describes the attribute feature associated with node $i$. To extend powerful convolutional neural network (CNN) to deal with graph-structured data, Bruna \textit{et al.} \cite{bruna2013spectral} model convolutional operation in the Fourier domain by computing the eigen-decomposition of the graph Laplacian, resulting in a heavy computation burden. Kipf \textit{et al.} \cite{kipf2016semi-supervised} propose graph convolutional networks (GCN) to simplify previous methods via a localized first-order approximation of spectral graph convolutions. Actually, the relations between objects are more complex in many applications. To encode high-order data correlation, Raman \textit{et al.} \cite{raman2017hypergraph} propose a hypergraph neural network by extending a simple graph to a hypergraph whose edge links more than two nodes. Recently, graph is used to deal with ZSL. Kato \textit{et al.} \cite{kato2018compositional} propose to compose classifiers for verb-noun pairs with GCN to recognize unseen human-object interactions. Wang \textit{et al.} \cite{wang2018zero-shot} use semantic embeddings of a category and knowledge graph encoding the relation of a novel category to familiar categories to predict object classes.

\subsection{Unseen Attribute-Object Composition Recognition}
The intuitive idea to recognize unseen attribute-object compositions is combining attribute classifiers and object classifiers \cite{chen2014inferring,misra2017from}. However, these classifier-based methods separately process the attributes and objects, ignoring the interdependent relations between the attributes and objects. As a result, these methods do not achieve satisfactory performance and suffer from the ``domain shift" problem \cite{fu2015transductive} - the distribution of the testing data is different from that of the training data. To explicitly factor out the attributes’ effect from their corresponding attribute-object composition representations, Nagarajan \textit{et al}. \cite{nagarajan2018attributes} propose to model an attribute as an \textit{operator} and an attribute-object composition as an object vector that is ``operated'' by the \textit{operator}. However, such linear and explicit matrix transformation may be insufficient to represent various attribute concepts, and Li \textit{et al.} \cite{li2020symmetry} propose a symmetry principle to learn the coupled and decoupled representations of attributes and objects in the group theory framework. Recently, generative models have been proven to be powerful in modeling data distributions. Nan \textit{et al.} \cite{nan2019recognizing} introduce a generative model with the encoder-decoder mechanism to reconstruct the input visual features. Our experiments further demonstrate that it is not which form of reconstruction loss (e.g., L2 loss used in \cite{nan2019recognizing}), but rather the reconstruction loss itself that helps learn generalized features (see Tab. \ref{tab:loss-functions}). Besides, Wei \textit{et al.} \cite{wei2019adversarial} propose to generate positive and (semi-) negative compositions to learn relations between the attributes and objects in an adversarial manner.

In this work, we resolve the unseen attribute-object recognition problem via modeling the relations between primitives through GCN, which is somewhat similar to the latest work CGE \cite{naeem2021learning} in network architecture. However, the design of our method is significantly different from CGE in many aspects. 1) Different graph nodes. In CGE, the graph nodes are composed of all attributes, objects, and compositions, which usually cause heavy overhead during the computation of graph convolutions. As a comparison, our graph nodes consist of only attributes and objects, which are often an order of magnitude lower in number than CGE. 2) Different graph structures. In CGE, two nodes are connected if they are co-occurring, which can be regarded as a special case of our link graph $\mathcal{G_L}$ (see Section \ref{sec_graph}). 3) Different loss function constraints. CGE uses cross-entropy loss based on dot product as the objective, while we change it to cross-entropy loss based on cosine logits, which is also adopted in \cite{mancini2021open} and shows a more stable optimization curve. Specifically, we have decoding loss and balance loss as two additional auxiliary loss functions, which are responsible for preserving important information \cite{nan2019recognizing} and alleviating bias in the generalized composition recognition \cite{purushwalkam2019task-driven}, respectively. A prior balance loss is proposed in \cite{xie2020region} that pursues the maximum similarity among seen and unseen compositions. However, our balance loss aims to raise the model's awareness of all unseen compositions.

\section{Multi-Attribute Dataset (MAD)}
In this section, we describe the motivation for constructing MAD, the process of collecting attribute-object annotations, and present basic statistics of MAD. In general, we aim to construct a dataset that truly reflects the multi-attribute nature of objects, thus applicable for testing unseen multi-attribute-object composition recognition algorithms.

\subsection{Motivation}
Currently, the two datasets (i.e., MIT-States and UT-Zappos50K) for unseen attribute-object composition recognition are all single-attribute annotations. In practice, only a single attribute is not enough to well describe a natural object. The single-attribute annotations are one-sided and easily cause ambiguity during inference (e.g., a \texttt{fresh sliced apple} is recognized as \texttt{fresh apple} but its label is \texttt{sliced apple}). Also, it is inaccurate to describe the co-occurrence patterns of the attributes, e.g., \texttt{old} and \texttt{rusty} for \texttt{car}. In addition, the two benchmarks suffer from label sparsity\cite{li2020symmetry} (some compositions with less than 5 images), noisy labels, and data insufficiency (with a maximum of 1,300 attribute-object compositions and 30k images, leading to fairly rapid over-fitting). Therefore, it is of great significance to construct a large-scale dataset annotated with multi-attribute.

\begin{figure*}[!t]
	\centering
	\includegraphics[width=1\textwidth]{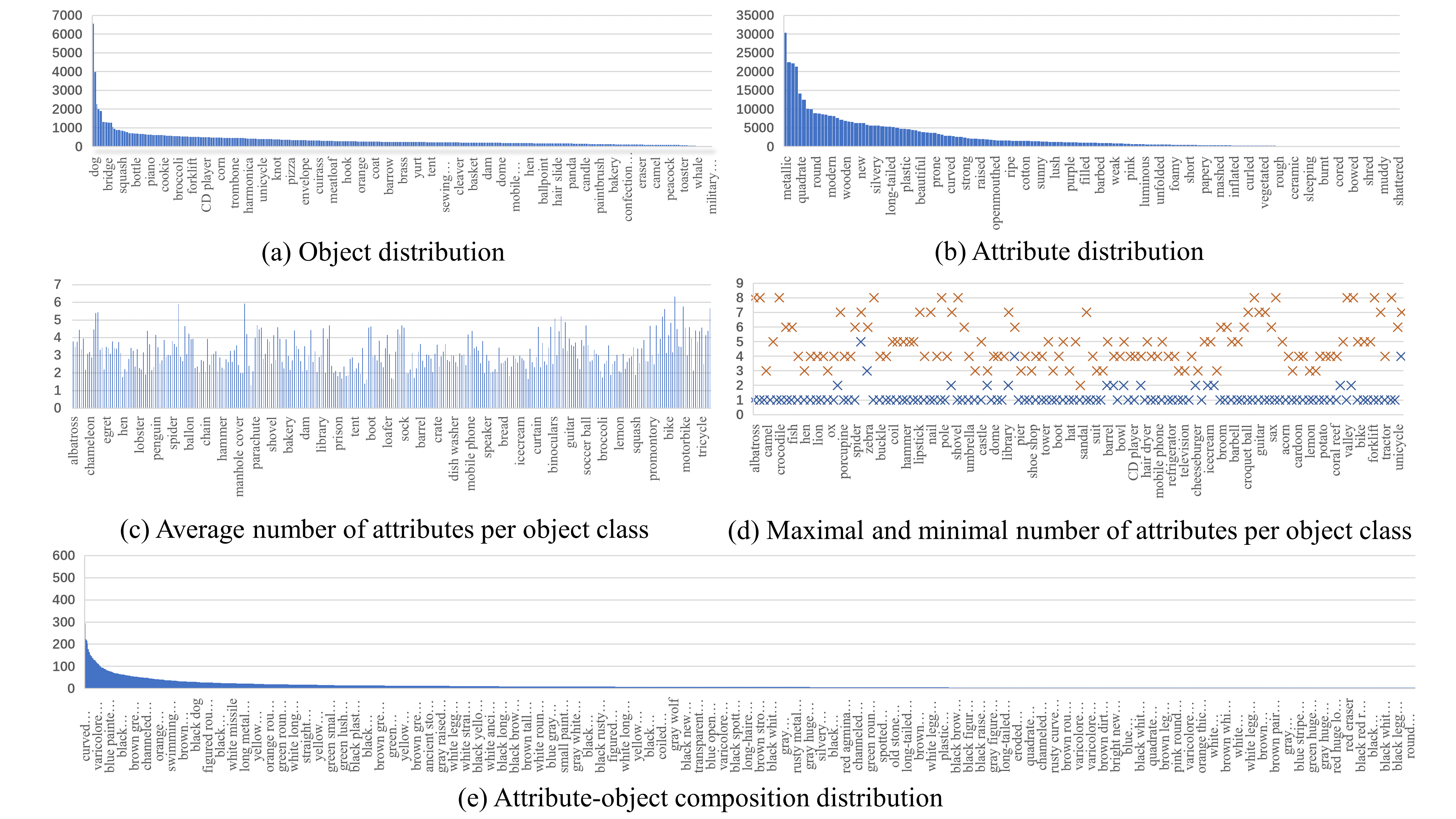}
	\caption{Statistical distributions of the MAD dataset.
	}
	\label{fig_mad} 
\end{figure*}

\subsection{Dataset construction}
\textbf{Step 1: collect images from ImageNet.} MAD is created based on the ImageNet \cite{deng2009imagenet:} dataset, which consists of object-centric images and provides the object labels. Therefore, we only focus on the annotation of the attribute part. We construct a sub-dataset by choosing 400 common object classes and randomly selecting 500 images from each class for the subsequent annotation process. Then, the images which satisfy one of the following three conditions are removed: 1) gray-scale (for color is a vital aspect of visual attributes), 2) with a prominent watermark, and 3) hard-to-recognize attributes.

\noindent \textbf{Step 2: determine the candidate list of objects and attributes.} It is of central importance to well determine the coverage of objects and attributes. The classes of objects come from the common objects in the ImageNet dataset. There are three sources for the attributes candidate list, 1) attributes that are used in papers related to attribute prediction task \cite{krishna2017visual,li2016richly,liu2015deep,zhao2019large}, 2) common attributes in daily life, and 3) complemented attributes that are manually added by previewing images to be annotated. After manually cleaning ambiguous/difficult to recognize attributes, we obtain a set of 158 unique attributes and 309 unique objects which are
used throughout the rest of the process. (A complete list of attributes and objects is given in Suppl Section C).

\noindent \textbf{Step 3: group attributes and objects and build an annotation application.} An object super-class (e.g., \texttt{animal}) generally shows limited attribute patterns. For example, attributes used to describe plants are not applicable to characterize animals. Therefore, objects are grouped together according to their super-class \cite{fan2019shifting,lin2014microsoft}. Besides, attributes are grouped by their super-class (e.g., \texttt{color}) to speed up the annotation process. Thus, annotators can first identify the super-class of attributes and then annotate specific attributes, which is otherwise difficult to label directly for hundreds of attributes. Then, we build an application to help annotate the raw data. In addition, it is observed that most instances of the same object class share many attributes, which might help improve annotation efficiency. For example, \texttt{clock} is usually \texttt{round} and we can treat \texttt{round} as a default attribute of the \texttt{clock} class without having to label it every time. Thus, the function of “pre-annotate” is added to the software to label common attributes of the object and any default attribute not matching the loaded image can be removed. 

\noindent \textbf{Step 4: annotate and review.} We do not require judgments about the existence of all attributes, which is impractical and extremely error-prone. Instead, annotators are required to label one to eight \textbf{salient} attributes for each image. Besides, images of the same object class are equally distributed among four annotators so as to reduce the risk of subjective bias towards the object. In total, 30 annotators are employed in the annotation project. Because attribute annotation is somewhat subjective, it may happen that some attributes are incorrectly labeled. Thus, another nine experienced workers are asked to review the preliminary annotation results and delete the attributes that are wrongly annotated.

\subsection{Statistical analysis}
The basic statistical properties of MAD are visualized in Figure.~\ref{fig_mad} and analyzed below.

Distribution of objects (Figure.~\ref{fig_mad}(a)). Most object classes contain hundreds of images, and the average number of images per object class is 375.7. Besides, there are some large object classes (11 objects) that contain thousands of images (e.g., \texttt{dog}) and some small object classes (19 objects) whose sizes are under 100 (e.g., \texttt{whale}).

Distribution of attributes (Figure.~\ref{fig_mad}(b)). For MAD, 8 different general categories of attributes are created (e.g., \texttt{color}, \texttt{pattern}, \texttt{shape}), which are further subdivided into 158 specific attributes (e.g., under the \texttt{pattern} category are \texttt{barbed}, \texttt{broken}…). For \texttt{containers}, their \texttt{shape} is one of the attributes that are described, while for \texttt{animals}, their state of \texttt{motion} is considered an important category. As a result, some attributes appear more frequently in the dataset, e.g., \texttt{metallic} appears 30,369 times, whereas others (17 attributes) only appear dozens of times, which poses severe data imbalance.

Average number of attributes per object class (Figure.~\ref{fig_mad}(c)). According to the statistics, the average number of attributes labeled for an object is closely related to the object itself, ranging from 1.69 for \texttt{window screen} to 6.44 for \texttt{harvester}, and the average number of attributes described per object is 3.33.

Maximal and minimal number of attributes per object class (Figure.~\ref{fig_mad}(d)). For different images, objects are annotated with different numbers of attributes, from as few as one attribute to eight attributes. The maximal and minimal number of attributes can be calculated among images with the same object class.

Distribution of attribute-object compositions (Figure.~\ref{fig_mad}(e)). Many images in the dataset share the same label (e.g., 490 images are labeled as \texttt{curved modern metallic bridge}). This commonality helps to capture the most common and essential attributes of an object.

\textbf{Compare with benchmark datasets.} There are two benchmark datasets (i.e., MIT-States and UT-Zappos50K) that are widely used in the unseen attribute-object composition recognition task. In terms of the size of dataset, MAD contains 158 attributes, 309 objects, and 8,030 compositions, a size much larger than other datasets. MAD details various attributes that are not included in the two other datasets, for example, the \texttt{motion} information (e.g., \texttt{standing}, \texttt{swimming}, \texttt{jumping}), and \texttt{distribution} information (e.g., \texttt{scattered}, \texttt{stacked}), which altogether increase the diversity of the dataset. On the other hand, the attribute number per object of MAD is not fixed, ranging from one to eight, which is more realistic and challenging. We split the dataset into a training set containing 81,371 images/5630 compositions, a validation set containing 27,104 images from 1,000 seen and 1,000 unseen compositions, and a testing set with 42,013 images from 1,400 seen and 1,400 unseen compositions. Tab.~\ref{tab:datasets-comparison} summarizes the main differences between these three datasets. For more details about comparisons between datasets please refer to Suppl Section C. Further, we propose two novel metrics suited for MAD, which will be detailed in Section \ref{sec_baselines_metrics}.

Compared to the single-attribute benchmarks, MAD poses three challenges that make it different from single-attribute datasets not only at the input/output level: 1) relation learning between the attributes in a composition, which does not need to be considered in the single-attribute case, 2) the more severe combination explosion problems due to the multiple attributes in a composition, and 3) the more significant long-tailed distribution of compositions. To address these challenges, we propose to use GCN to tackle single-/multi-attribute composition recognition at the same time, where the nodes of the graph are composed of attributes and objects. GCN can not only learn the relationships between the objects and attributes (which needs to be considered in the single-attribute setting), but also model associations between attributes. Furthermore, as long as the number of attributes and objects remains the same, no matter how many attributes are included in each composition, the number of nodes in the graph network model remains the same and it is computationally affordable. The third challenge is not our main focus in this paper, and we will address it in future work.

\section{Method}
\subsection{Problem Formulation}
For unseen attribute-object composition recognition, the training set (seen compositions) is defined as $\mathcal{S}={\{x_i^s, y_i^s\}}_{i=1}^{n_s}$, where $x_i^s$ is the $i$-th training image with ${(a_i^1, a_i^2,..., a_i^{l_i}, o_i)}^s$ as its attribute-object composition label, $y_i^s \in \mathcal{Y}^s=\{1, 2,..., |c^s|\}$ denoting the seen image class. $l_i$ is the number of attributes labeled in the $i$-th image and is greater than 1 when adapted to the multi-attribute setting. The testing set (unseen compositions) is defined as $\mathcal{U}={\{x_j^u,y_j^u\}}_{j=1}^{n_u}$, where $x_j^u$ is the $j$-th testing image with ${(a_j^1, a_j^2,..., a_j^{l_j},o_j)}^u$ as its attribute-object composition label, $y_j^u \in \mathcal{Y}^u=\{|c^s|+1, |c^s|+2,..., |c^s|+|c^u|\}$ denoting the unseen image class. The seen composition set and unseen composition set satisfy $\mathcal{Y}^s\cap \mathcal{Y}^u=\emptyset$, $\mathcal{Y}^s  \cup \mathcal{Y}^u=\mathcal{Y}$. In the following text, we will omit the superscript $s/u$ to make the expression simple without causing confusion.

\begin{figure*}[!t]
	\centering
	\includegraphics[width=1\textwidth]{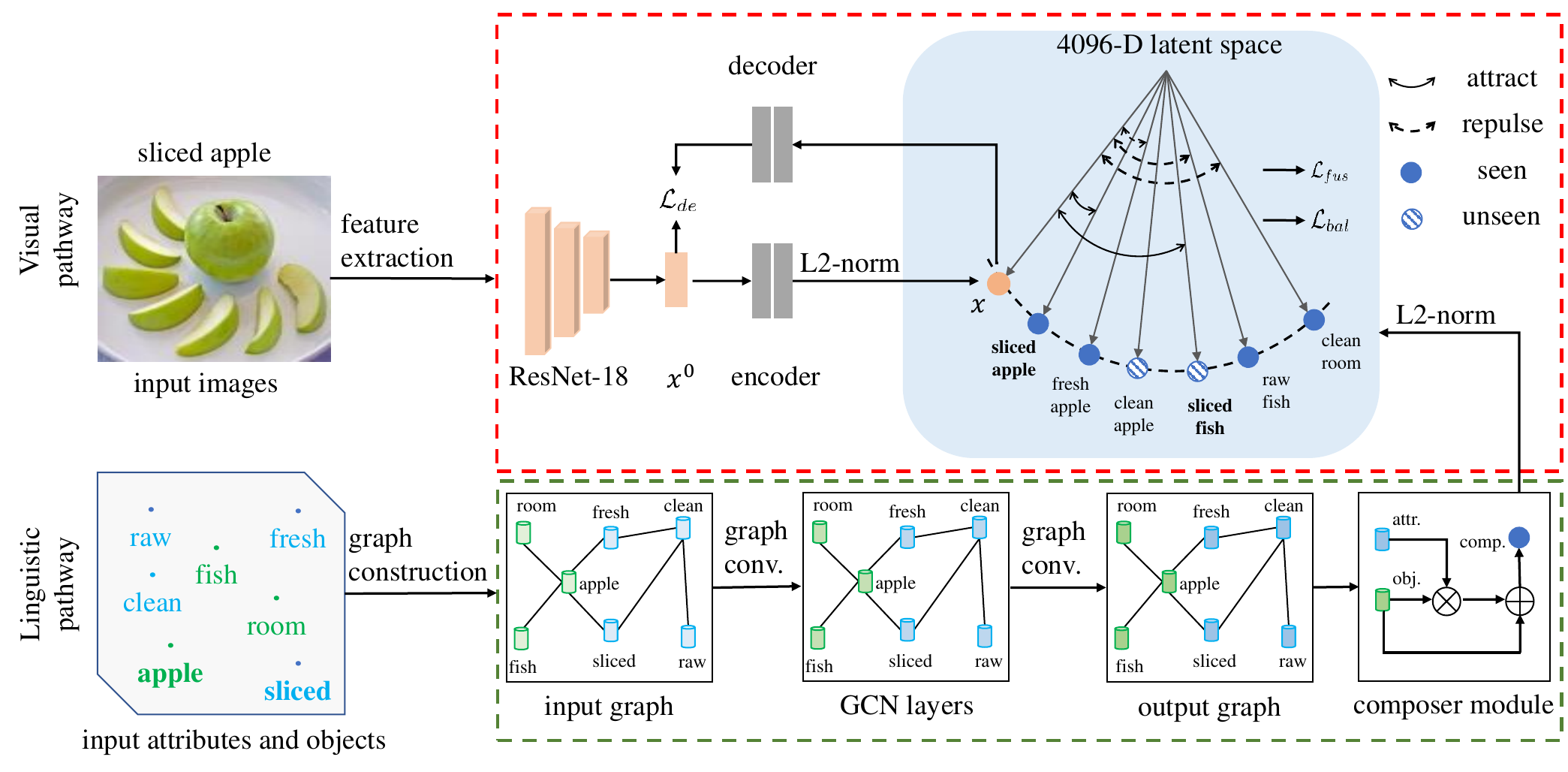}
	\caption{The overview of our network architecture. The network is composed of the visual pathway and the linguistic pathway. In the visual pathway, an input image is processed by the pre-trained ResNet-18, the visual encoder module and the L2 normalization layer, obtaining the encoded visual feature $x$. In the linguistic pathway, attribute(s) and object corresponding to the input image are embedded and processed by the linguistic encoder module implemented by graph convolutional networks (GCN). After that, a composer module fuses the attribute(s) and object node representations from the output graph of GCN, obtaining the linguistic encoded feature $z$, which is then L2 normalized. The model is optimized by minimizing the contrastive form based fusion loss $\mathcal{L}_{fus}$ between the two-modal data, as well as two other regularizers $\mathcal{L}_{bal}$ and $\mathcal{L}_{de}$ to learn discriminative features. During the testing, the linguistic encoded features of all possible attribute-object compositions are computed and compared with the encoded visual feature of the input image to predict the most likely attribute-object composition.
	}
	\label{fig_framework}
\end{figure*}

\subsection{Method Overview}
As shown in Figure.~\ref{fig_framework}, we propose a two-pathway network architecture consisting of a visual pathway to process visual data and a linguistic pathway to process linguistic data. In the visual pathway, given a training image, a pre-trained convolutional neural network (e.g., ResNet-18) is used to extract initial visual feature $x^{0} \in \mathbb{R}^m$, which is further processed by the visual encoder module $f_{en}(\cdot)$, obtaining encoded visual feature $x\in\mathbb{R}^k$ in a latent space. In the linguistic pathway, the attribute and object labels corresponding to the input image are embedded as $z_a$ and $z_o$ respectively via graph convolution operations. After that, a composer module is responsible for fusing attribute-object information, outputting encoded linguistic feature $z \in \mathbb{R}^k$ in the latent space.

To reduce mis-classifications caused by similar compositions, we optimize the contrastive loss \cite{he2020momentum,pmlr-v119-chen20j} to learn discriminative visual features. By doing this, the input image fits the correct linguistic label whilst repels incorrect linguistic labels. During the testing, we identify the attribute-object composition prediction of a testing image by searching for the composition that has the highest cosine similarity to the encoded visual feature of the given testing image.

\subsection{Network Architecture and Data Flow}
\subsubsection{Visual Encoder}
Given an input training image $I$, we use a pre-trained convolutional neural network to extract $m$-dimensional image features, obtaining the initial image features $x^0\in\mathbb{R}^m$. Then, we project $x^0$ to a $k$-dimensional latent space using a visual encoder module $f_{en}(\cdot)$, which is implemented by a simple one-layer fully connected network with a non-linear activation function Softplus. The transformation is defined as follows:
\begin{equation}
x=f_{en}\left(x^0;\phi\right),
\label{eq_visualEn}
\end{equation}
where $\phi$ represents the learnable parameters in the visual encoder $f_{en}(\cdot)$. 
Further, we enforce $x$ to be unit-length via a L2-normalization layer: $x=\frac x {\left\|x \right\|_{2}}$.

\subsubsection{Visual Decoder}
Inspired by the auto-encoder architecture used in recent works \cite{kodirov2017semantic,nan2019recognizing}, we design the visual decoder function $f_{de}(\cdot)$ projecting $x$ as $\hat{x}$ to reconstruct the initial image feature $x^0$:
\begin{equation}
\hat{x}=f_{de}\left(x;\theta\right),
\label{eq_de}
\end{equation}
where $\theta$ represents the learnable parameters in the decoder $f_{de}(\cdot)$. For implementation, we choose a linear decoder function since it work well in practice. The motivation for designing the visual decoder is to learn the intrinsic representation of $x$ by applying a strong constraint to the latent space, making the encoded visual feature preserve important information in $x^0$.

\subsubsection{Linguistic Encoder}\label{sec_graph}
The recognition of composition is inseparable from the learning of primitives. In turn, good representations of primitives will facilitate the learning of advanced compositions. Therefore, it is natural to consider utilizing a graph framework with a node representing an attribute/object to learn good primitive embeddings. Besides, the relations between primitives including attribute-attribute, object-object and attribute-object are complex, such as \texttt{folded box}/\texttt{unfolded box} (antonym attributes) and \texttt{traffic light}/\texttt{traffic sign} (high co-occurrence objects). To encode these complex relations as well as learn embeddings, we use GCN to aggregate and propagate node information, in which each graph node represents an attribute or an object. GCN is a multilayer neural network with a convolution operation defined on a graph. The input of GCN consists of a graph adjacency matrix and a node feature matrix. We initialize the graph feature matrix $Z^0\in\mathbb{R}^{N\times d}$ by extracting the $N$ attributes and object feature vectors with pre-trained word embedding models, where $d$ is the dimensionality of the word embedding vectors. By default, we use the Glove \cite{pennington2014glove:} model to extract the linguistic features.  

\textbf{Graph structure.} The next key is to determine the adjacency matrix $A\in\mathbb{R}^{N\times N}$, in which $A(i,j)$ indicates the semantic correlation strength between node $i$ and $j$. In this paper, we construct four different kinds of attribute-object semantic association graphs to explore a reasonable graph structure to model the relations between the attributes and objects. The four graphs are vanilla random graph, sparse random graph, link graph and embedding graph, respectively, which are detailed as bellow. 

\noindent (1) Vanilla random graph $\mathcal{G_{R}}$: the adjacency matrix is randomly initialized with a normal distribution and set to be learnable, i.e., $A(i,j) \sim \mathcal{N}(0, 0.01)$. 

\noindent (2) Sparse random graph $\mathcal{G_{SR}}$: it is worth noting that not all attribute-object compositions are linguistically and physically meaningful in the real world, such as \texttt{flying dog} and \texttt{sliced boat}. To disenable unreasonable connections, we regularize the randomly initialized adjacency matrix with the L1 norm constraint to make the graph sparse. 

\noindent (3) Link graph $\mathcal{G_L}$: it seems reasonable that two nodes can be linked if the two nodes correspond to an existing composition in the training set. Taking the complicated multi-attribute setting as an example, the node representing \texttt{brown} is linked to the node \texttt{snake}, and the node representing \texttt{striped} is linked to the node \texttt{swimming} and so on when the composition \texttt{brown striped swimming snake} is in the training set. Formally, it is expressed as:
\begin{equation}
	A(i,j)=\begin{cases}
		\ 1,  \ & \textnormal{if \ } \textnormal{ComIndex} (i, j) \in \mathcal{Y}^s, \\
		\ 0,   \ & \textnormal{else}. \\
	\end{cases}
	\label{eq_link_graph}
\end{equation}
where $\textnormal{ComIndex}(i,j)$ is the composition index of node $i$ and $j$ in the training label set $\mathcal{Y}^s$. 

\noindent (4) Embedding graph $\mathcal{G_E}$: since we encode attribute and object labels with word embeddings, it is natural to consider that two nodes are connected if the distance between their word embeddings is lower than a threshold. Motivated by \cite{henaff2015deep}, we compute $A\left(i,j\right)$ of the embedding graph by a Gaussian diffusion kernel:
\begin{equation}
A(i, j)=\exp \left(-\frac{\left\|z_{i}^{0}-z_{j}^{0}\right\|_{2}^{2}}{s^2}\right),
\label{eq_embed}
\end{equation}
where $z_{i}^0$ and $z_{j}^0$ are the $i$-th and $j$-th rows of $Z^0$, and $s$ is a hyper-parameter. For simplicity, it is set to 1. However, due to the inaccurate semantic correlations caused by the introduction of the word embedding model, there could be some noisy connections. Thus, we binarize $A$ in Eq.~\ref{eq_embed} to be a 0-1 matrix to filter out noisy connections:
\begin{equation}
A=\mathcal{I}_{A\left(i,j\right)>t},
\end{equation}
where $\mathcal{I}$ is the indicator function, outputting 1 if $A\left(i,j\right)>t$ holds and 0 otherwise. $t$ is a threshold value, which is set to 0.9. A larger threshold means a sparser graph. In the experiments (Section \ref{sec_graph_relation}), we observe that the sparse random graph consistently performs better than other graphs. Therefore, we use the sparse random graph throughout the paper if not specified.

\textbf{Node representation updating.} The following layer takes the previous layer feature matrix $Z^l\in\mathbb{R}^{N\times d_l}$ and the adjacency matrix $A$ (where $d_l$ denotes the dimensionality of node representations in the $l$-th layer of GCN) as inputs, and it outputs the updated node representations $Z^{l+1}\in\mathbb{R}^{N\times d_{l+1}}$. Specifically, the propagation rule \cite{kipf2016semi-supervised} of the $l$-th layer is defined as:
\begin{equation}
Z^{l+1}=f\left(\hat{A}Z^lW^l\right),
\end{equation}
where $\hat{A}$ is the normalized adjacency matrix, $W^l$ is layer-specific learnable parameters and $f(\cdot)$ represents the non-linear activation function ReLU. For the last layer, the output of the GCN is $Z\in\mathbb{R}^{N\times k}$, where $k$ denotes the dimensionality of the common latent space. 

\subsubsection{Attribute-Object Composer}
The GCN is responsible for learning good node-level representations via modeling complex node correlations, but not good at generating an integrated composition-level representation. In general, the attribute and object are two fundamentally different entities with the attribute more abstract. Therefore, we propose a simple attention-based composer module to fuse attribute and object representations instead of adding them together directly. Given the output graph node representation matrix $Z$, we extract the attribute node representations ${z_a^1,z_a^2,...,z_a^l }$ and object node representation $z_o$ corresponding to the input image ($l$ is the number of attributes of the image). The fusion transformation is defined as:
\begin{equation}
	z=\sigma \left(\sum_{i=1}^l W^{cp}z_a^i\right) \otimes z_o + z_o,
	\label{eq_composer}
\end{equation}
where $\sigma(\cdot)$ is the sigmoid function, $W^{cp} \in \mathbb{R}^{k \times k}$ is a learnable weight matrix in the composer module and $\otimes$ is the Hadamard product that carries out product operation element-wise. In fact, the composer module can accept input with an arbitrary number of attributes, and attributes and objects not present in the input image are not accounted for in Eq. \ref{eq_composer}, i.e., they are masked out. Thus, our model can be adapted to both single-/multi-attribute recognition tasks, where the composer module outputs a holistic composition feature. Lastly, we normalize $z$ to be unit-length: $z=\frac z {\left\|z \right\|_{2}}$.

\subsection{Loss Functions}
\subsubsection{Fusion Loss}
The goal of fusion loss is to learn a high compatibility score between the encoded visual feature $x$ and the correct linguistic feature $z$. One common used objective function is the L2 loss (cosine loss) \cite{nan2019recognizing} or triplet loss\cite{wei2019adversarial}. However, the L2 loss or triplet loss minimizes the distance between $x$ and $z$ and can only maximize the distance between $x$ and the finite negative compositions (0 for the L2 loss and 1 for the triplet loss). In other words, they cannot minimize the distance between $x$ and $z$ and maximize the distance between $x$ and all the remaining negative compositions at the same time, resulting in less discriminative features. Inspired by the alignment (closeness of features from positive pairs) and uniformity two key properties of feature distribution induced by contrastive loss\cite{he2020momentum,pmlr-v119-chen20j,wang2020understanding}, we utilize it to measure the difference between $x$ and $z$:

\begin{equation}
\mathcal{L}_{fus}=- \frac 1 b \sum_{i=1}^{b} \log \frac {\exp (\tau \cdot {\rm sim}(x_i, z_{y_i}))} {\sum_{j \in \mathcal{Y}^s}  \exp (\tau \cdot {\rm sim}(x_i, z_j))},
\label{eq_fus} 
\end{equation}
where $b$ is the batch-size and $\tau$ is a temperature hyper-parameter that controls the concentration of feature distribution on our unit sphere, and ${\rm sim}(x_i, z_j)$ is the cosine similarity function: ${\rm sim}(x_i, z_j)=\frac {x_i^\mathrm{T} z_j} {{\left\|x_i \right\|_{2}}{\left\|z_j \right\|_{2}}}={x_i^\mathrm{T} z_j}$. Note that both $x_i$ and $z_j$ are L2 normalized. Intuitively, minimizing the fusion loss will assign high scores to the correct compositions and low scores to the remaining all incorrect ones, which leads to optimization behavior different from the L2 loss and triplet loss. Therefore, similar compositions are expected to be separated or have a large margin in the latent space, which is helpful to reduce mis-classifications.

\subsubsection{Balance Loss}
To tackle the challenge of domain bias in generalized unseen attribute-object composition recognition \cite{purushwalkam2019task-driven}, we propose a simple balance loss by maximizing the minimum of similarity among $x$ and unseen compositions $z_k \in \mathcal{Y}^u$, which is formulated as:
\begin{equation}
\mathcal{L}_{bal}=-\frac 1 b \sum_{i=1}^b \min \limits _{k \in \mathcal{Y}^u} {\rm sim}(x_i, z_k).
\end{equation}
Intuitively, minimizing the balancing loss will enforce the encoded visual feature $x$ to approach the least similar unseen composition, which helps raise the awareness of all unseen compositions during the training stage. Another formulation of balance loss is proposed in \cite{xie2020region} (denoted as $\mathcal{L}_{bal}^*$, see Tab.~\ref{tab:loss-functions}), which pursues the maximum similarity consistency among seen and unseen compositions. However, we argue that aligning the maximum similarity among seen and unseen compositions cannot benefit reducing bias in the generalized scenario as the less similar unseen compositions would not be likely taken as the candidate solutions.

\subsubsection{Decoding Loss}
Inspired by recent auto-encoder works \cite{nan2019recognizing, kodirov2017semantic} on unseen attribute-object composition recognition, we introduce the decoding loss to minimize the L2 loss between the initial visual feature $x^0$ and the reconstruction visual feature $\hat{x}$:
\begin{equation}
\mathcal{L}_{d e}=\frac 1 b \sum_{i=1}^b \left\|x_i^0-\hat{x}_i\right\|_{2}.
\end{equation}
It is expected to preserve important information in input visual features via minimizing the decoding loss.

\subsection{Learning and Inference}
During the training, we input the initial visual features $x^0$ extracted from pre-trained convolutional networks and graph feature matrix $Z^0$ initialized with word embeddings. Note that attributes and objects not present in the image are masked out to adapt either the single-attribute or the multi-attribute setting (see Eq. \ref{eq_composer}). The goal is to learn the optimal parameters that minimize the overall loss:
\begin{equation}
L\left(W\right)= \mathcal{L}_{fus}+\beta \mathcal{L}_{bal}+\gamma \mathcal{L}_{de},
\label{eq_loss}
\end{equation}
where $W$ represents all of the learnable parameters in the network. We use ADAM \cite{kingma2014adam:} as our optimizer.


The inference is carried out in the generalized setting\cite{purushwalkam2019task-driven}, where it is assumed which single-/multi-attribute compositions are known to be present in the dataset as seen/unseen classes. Firstly, the graph node representation matrix $Z\in\mathbb{R}^{N\times k}$ is pre-computed and stored. At the same time, we construct a mask matrix $M \in \{0,1\}^{(|c^s|+|c^u|)\times N}$ with each row indicating the absolute positions of attributes and objects of a composition label. For example, suppose the attribute set is $\{a_1,a_2,a_3\}$ and the object set is $\{o_1,o_2,o_3\}$, then the mask matrix $M$ for $\{a_1,a_2,o_3\}$ is:
\begin{equation}
	[\underbrace{1}_{a_1} \quad \underbrace{1}_{a_2} \quad \underbrace{0}_{a_3} \quad | \quad \underbrace{0}_{o_1} \quad \underbrace{0}_{o_2} \quad \underbrace{1}_{o_3}].
\end{equation}

Then, holistic features $\tilde{Z}\in \mathbb{R}^{(|c^s|+|c^u|)\times k}$ for all compositions present in $\mathcal{Y}^s \cup \mathcal{Y}^u$ are computed by following Eq.~\ref{eq_composer}:
\begin{equation}
	\tilde{Z}=\sigma \left(M_{[:,:N_a]}Z_{[:N_a]}W^{cp}\right) \otimes \left(M_{[:, N_a:]}Z_{[N_a:]}\right) + M_{[:, N_a:]}Z_{[N_a:]},
\end{equation}
where $N_a$ is the cardinality of the attribute set. Note that Eq.~\ref{eq_composer} is expressed in a matrix form here. Given the testing image, we compute its encoded visual feature $x$, and derive a prediction by searching the composition label with the largest cosine similarity score:  
\begin{equation}
	\mathop{\arg\max}_{k \in {\mathcal{Y}^s \cup \mathcal{Y}^u}} {\rm sim}(x, \tilde{z}_k).
\end{equation}


\newcommand{\tabincell}[2]{\begin{tabular}{@{}#1@{}}#2\end{tabular}}

\section{Experiments}
\subsection{Datasets and Metrics}\label{sec_baselines_metrics}
\textbf{Benchmarks.} We evaluate the proposed model on two publicly available single-attribute datasets MIT-States\cite{isola2015discovering} and UT-Zappos50K \cite{yu2014fine-grained}, and the multi-attribute dataset MAD. The MIT-States dataset is composed of 63,440 images, covering 115 attribute classes, 245 object classes, and 1,962 attribute-object compositions. Each image is annotated with an attribute-object composition label such as \texttt{clean kitchen}. The same settings are used as in previous works \cite{purushwalkam2019task-driven,li2020symmetry,naeem2021learning}: 1,262 compositions/30,338 images are used for the training and 800 compositions/12,995 images for the testing. UT-Zappos50K is a fine-grained shoes dataset with 50,025 images, including 16 attribute classes, 12 object classes, and 116 attribute-object compositions. Following the settings in previous works \cite{purushwalkam2019task-driven,li2020symmetry,naeem2021learning}, we use 83 compositions/22,998 images for the training and 36 compositions/2,914 images for the testing. Detailed statistics are shown in Tab.~\ref{tab:datasets-comparison}.

\begin{table*}[!t]
	\centering
	\caption{Dataset statistics. a: \# attributes, o: \# objects, p: \# compositions, a/i: \# attributes per image, sp: \# seen compositions, up: \# unseen compositions, i: \# images.}
	\label{tab:datasets-comparison}
	\setlength
	\tabcolsep{7pt}
	\large
	\renewcommand\arraystretch{1.05}{
		\resizebox{0.96\textwidth}{!}{
			\begin{tabular}{lcccc|cc|ccc|ccc}
				& & & & &\multicolumn{2}{c}{Training} &\multicolumn{3}{c}{Validation} &\multicolumn{3}{c}{Testing}  \\
				Datasets &a &o &p &a/i &sp &i &sp &up &i &sp &up &i \\
				\hline
				MIT-States\cite{isola2015discovering} &115 &245 &1,962 &1 &1,262 &30,338  &300 &300  &10,420  &400 &400 &12,995\\
				UT-Zappos50K\cite{yu2014fine-grained} &16 &12 &116 &1 &83 &22,998  &15  &15  &3,214  &18 &18 &2,914\\
				MAD &158 &309 &8,030 &1-8 &5,630 &81,371 &1,000 &1,000  &27,104  &1400 &1400 &42,013\\
				
	\end{tabular}}}
\end{table*}

\textbf{Evaluation metrics.} In the single-attribute case, it is reasonable to assess the performance of a model by simply comparing whether the predicted attribute-object composition matches the ground truth label. However, in the multi-attribute scenario, it has not been previously explored how to evaluate the results where some of the predicted attributes are the same as the ground truth but not the same as the rest. For example, there are two models A and B available and there is a dog image labeled \texttt{black long-haired standing strong}. Model A predicts \texttt{black strong} and model B predicts \texttt{black long-haired standing}. Although they both give false predictions (evaluated under the single-attribute case), intuitively, B performs better than A because three-quarters of the attributes predicted by B are correct. Thus, we use the single-/multi-attribute evaluation metrics to give a comprehensive performance evaluation.

\textbf{Single-attribute evaluation metrics.} For the single-attribute setting, we closely follow the state-of-the-art methods \cite{li2020symmetry,naeem2021learning} and report area under the curve (AUC) \cite{purushwalkam2019task-driven} under the generalized unseen attribute-object recognition setup. As the unseen compositions are not involved in the training, the model tends to be biased toward seen compositions. Therefore, to make a trade-off between the accuracy of seen and unseen compositions, a varying scalar term is added to unseen compositions, and AUC is computed by estimating the area under the unseen-seen accuracy curve. Besides, we report the best Seen (tested on seen compositions), Unseen (tested on unseen compositions), and HM (harmonic mean of Seen and Unseen) metrics.

\begin{figure}[!h]
	\centering
	\includegraphics[width=0.48\textwidth]{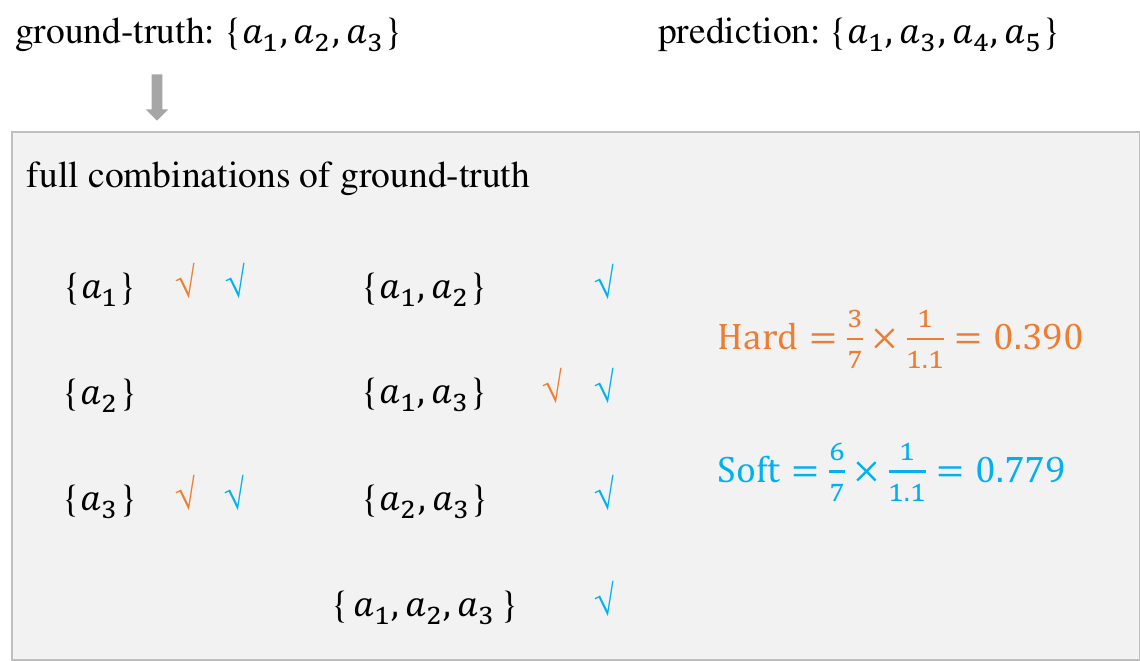}
	\caption{The illustration of the computation process of Hard and Soft. For convenience, the object is assumed to be classified correctly.
	}
	\label{fig_metric}
\end{figure}

\textbf{Multi-attribute evaluation metrics.}
To take the presence of multiple attributes into consideration, one may seek the metrics commonly used in multi-label classification \cite{chen2019multi-label,chen2019learning}, including the overall precision, recall, F1-measure (OP, OR, OF1), per-class precision, recall, F1-measure (CP, CR, CF1) and mean average precision (mAP). However, a potential assumption of these metrics is that the probability of occurrence of each label is assumed to be the same. For the multi-attribute setting, while an image contains a single object, it might contain multiple attributes, resulting in attributes and objects that are not of equal importance. Therefore, these evaluation metrics would favor methods better predicting attributes even if they do not classify the object correctly. From the perspective of compositionality, we propose two novel metrics, Hard and Soft, which are defined below:
\begin{align}
	\textnormal{Hard} &=\frac 1 n \sum_{i=1}^n \frac {\sum_{j \subset {\rm comb}({gt}_a^i)}\mathcal{I}_{j \bm{\subset} {pre}_a^i}} {\# {\rm comb}({gt}_a^i)} \cdot \mathcal{I}_{{gt}_o^i={pre}_o^i} \cdot C, \label{eq_hard} \\
	\textnormal{Soft} &=\frac 1 n \sum_{i=1}^n \frac {\sum_{j \subset {\rm comb}({gt}_a^i)}\mathcal{I}_{j \bm{\cap} {pre}_a^i}} {\# {\rm comb}({gt}_a^i)} \cdot \mathcal{I}_{{gt}_o^i={pre}_o^i} \cdot C, \label{eq_soft} \\
	C &= \begin{cases}
		\frac 1 {\eta^{{\rm len}({pre}_a^i)-{\rm len}({gt}_a^i)}}, & \textnormal{if \ } \; {\rm len}({pre}_a^i) \textgreater {\rm len}({gt}_a^i)\\
		1, & \textnormal{else}
	\end{cases} \label{eq_calibration}
\end{align}
where $n$ is the number of images, ${\rm comb}(\cdot)$ is the combination function that returns full combinations for input sequence, $\mathcal{I}$ is the indicator function, ${gt}_a^i$ and ${pre}_a^i$ are the ground-truth and predictions of the attributes of the $i$-th image, respectively, $C$ is a calibration term and $\eta$ is a calibration factor (set to 1.1). There are three key points in the formulas. 1) Hard and Soft are computed by separately considering the contribution of attributes and objects, and eventually, their effects are balanced. 2) By traversing the full combinations of ground-truth and calculating the degree of correlation ($\subset$ in Eq. \ref{eq_hard} and $\cap$ in Eq. \ref{eq_soft}) between predictions and the full combinations, the new metrics emphasize the compositional nature of the problem itself, rather than focusing on whether a composition is seen or unseen like AUC. 3) The metrics will be calibrated if the prediction length exceeds the length of ground-truth. To make the computation process clear, we provide a toy example, as illustrated in Figure. \ref{fig_metric}. Suppose the ground-truth of the attributes of the image is $\{a_1, a_2,a_3\}$, and the prediction is $\{a_1, a_3, a_4, a_5\}$. According to the definitions, the value of Hard and Soft are 0.390 and 0.779, respectively. Generally, Hard is lower than Soft for "$\subset$" is more difficult to satisfy than "$\cap$" between two sets.

\begin{table*}[!t]
	\centering
	\caption{Comparison results on AUC \cite{purushwalkam2019task-driven} metric on the MIT-States and UT-Zappos50K datasets (\%). We report the top-k (k=1,2,3) AUC on the validation/testing set, and the best Seen, Unseen and HM (harmonic mean) on the testing set. All methods use ResNet-18 as the backbone.}
	\label{tab:methods-comparison-single}
	\setlength
	\tabcolsep{7pt}
	\large
	\renewcommand\arraystretch{1.2}{
		\resizebox{1\textwidth}{!}{
			\begin{tabular}{lccccccp{0.7cm}<{\centering}p{1.1cm}<{\centering}p{0.8cm}<{\centering}ccccccp{0.7cm}<{\centering}p{1.1cm}<{\centering}p{0.8cm}<{\centering}}
				\toprule
				\multirow{3}{*}{Methods} &\multicolumn{9}{c}{MIT-States} &\multicolumn{9}{c}{UT-Zappos50K}  \\
				\cmidrule(lr){2-10}  \cmidrule(lr){11-19}
				&\multicolumn{3}{c}{Val AUC} & \multicolumn{3}{c}{Test AUC} 
				&\multirow{2}{*}{Seen}  &\multirow{2}{*}{Unseen}  &\multirow{2}{*}{HM}  
				&\multicolumn{3}{c}{Val AUC} & \multicolumn{3}{c}{Test AUC} 
				&\multirow{2}{*}{Seen}  &\multirow{2}{*}{Unseen}  &\multirow{2}{*}{HM} \\
				\cmidrule(lr){2-4}  \cmidrule(lr){5-7}  \cmidrule(lr){11-13}  \cmidrule(lr){14-16}  
				& 1 & 2 &3 & 1 & 2 &3  & & &  & 1 & 2 &3 & 1 & 2 &3\\ 
				\cmidrule(lr){1-1} \cmidrule(lr){2-4}  \cmidrule(lr){5-7}  \cmidrule(lr){8-10} \cmidrule(lr){11-13}  \cmidrule(lr){14-16} \cmidrule(lr){17-19}
				AttOperator \cite{nagarajan2018attributes} &2.5 &6.2 &10.1 &1.6 &4.7 &7.6  &14.3 &17.4  &9.9  &21.5 &44.2 &61.6 &25.9 &51.3 &67.6  &59.8 &54.2  &40.8\\
				RedWine \cite{misra2017from} &2.9 &7.3 &11.8 &2.4 &5.7 &9.3  &20.7  &17.9  &11.6  &30.4 &52.2 &63.5 &27.1 &54.6 &68.8  &57.3  &62.3  &41.0\\
				LabelEmbed+ \cite{nagarajan2018attributes} &3.0 &7.6 &12.2 &2.0 &5.6 &9.4  &15.0 &20.1  &10.7  &26.4 &49.0 &66.1 &25.7 &52.1 &67.8  &53.0 &61.9  &40.6\\
				GenModel \cite{nan2019recognizing} &3.1 &6.9 &10.5 &2.3 &5.7 &8.8  &24.8 &13.4  &11.2  &31.3 &49.6 &68.7 &19.2 &37.3 &50.3  &47.2 &51.2  &37.1\\
				TMN \cite{purushwalkam2019task-driven} &3.5 &8.1 &12.4 &2.9 &7.1 &11.5 &20.2  &20.1  &13.0  &36.8 &57.1 &69.2 &29.3 &55.3 &69.8 &58.7  &60.0  &45.0\\
				SymNet \cite{li2020symmetry} &4.3 &9.8 &14.8 &3.0 &7.6 &12.3  &24.4  &25.2  &16.1    &25.9 &50.9 &64.5 &23.9 &48.2 &64.4  &53.3  &57.9  &39.2\\
				CompCos \cite{mancini2021open} &5.9 &13.3 &19.4 &4.5 &10.4 &16.3 &25.3 &24.6 &16.4 &37.8 &60.0 &73.0 &28.7 &54.2 &71.5 &59.8 &62.5 &43.1 \\
				CGE \cite{naeem2021learning} &\textbf{6.8} &14.6 &20.2 &5.1 &11.8 &17.3 &28.7 &25.3 &17.2 &\textbf{38.7} &\textbf{60.2} &\textbf{73.2} &26.4 &55.6 &71.0 &56.8 &63.6 &41.2 \\
				\cmidrule(lr){1-1} \cmidrule(lr){2-4}  \cmidrule(lr){5-7} \cmidrule(lr){8-10} \cmidrule(lr){11-13}  \cmidrule(lr){14-16} \cmidrule(lr){17-19}
				Ours &\textbf{6.8}  &\textbf{14.8} &\textbf{21.6} &\textbf{5.5} &\textbf{12.6} &\textbf{18.5} &\textbf{30.0} &\textbf{26.1}  &\textbf{18.1}  &27.7  &54.8 &70.6 &\textbf{33.5} &\textbf{58.3} &\textbf{72.6} &\textbf{60.9} &\textbf{63.7}  &\textbf{49.4}\\ 
				\bottomrule
	\end{tabular}}}
\end{table*}

\subsection{Baselines} 
Eight baseline methods are compared with our method. We briefly introduce the baselines as follows:


\noindent \textemdash AttOperator \cite{nagarajan2018attributes} takes the attribute as an \textit{operator} and the attribute-object composition is modeled as the object vectors transformed by the attribute \textit{operator};

\noindent \textemdash RedWine \cite{misra2017from} uses pre-trained linear SVM weights to replace the word vectors and trains a neural network to recognize unseen attribute-object compositions;

\noindent \textemdash LabelEmbed+ \cite{nagarajan2018attributes} embeds the attribute, object, and image features into a joint feature space;


\noindent \textemdash GenModel \cite{nan2019recognizing} adopts the auto-encoder architecture that reconstructs the input image features to recognize unseen compositions;


\noindent \textemdash TMN \cite{purushwalkam2019task-driven} leverages the idea similar to "divide and conquer" to predict unseen compositions in a semantic concept space by constructing a set of light-weight neural network modules and configuring them via a gating function in a task-driven way.

\noindent \textemdash SymNet \cite{li2020symmetry} adopts the symmetry principle in attribute-object transformation to predict unseen compositions by decoupling the input features into separate attribute and object and then coupling them, which is optimized under the framework of group theory.

\noindent \textemdash CompCos \cite{mancini2021open} uses cosine similarity to estimate the feasibility score of each composition and exploits the scores to remove potential distractors in the embedding space.

\noindent \textemdash CGE \cite{naeem2021learning} models the dependency between attributes, objects, and their compositions within a graph network and predicts unseen compositions by learning a compatibility function that assigns high scores to matched visual-linguistic feature pairs.

Note that these baselines are originally proposed to address single-attribute-object composition recognition, and they accept only a \emph{single} attribute as input. To adapt to the multi-attribute situation, we simply average the word embeddings of multiple attributes in a composition label to keep the least modifications in model architecture. For example, we use $\frac {{\rm emb}(\texttt{fresh}) + {\rm emb}(\texttt{sliced})} 2$ as a \emph{single} linguistic attribute input for a \texttt{fresh sliced apple} composition label, where ${\rm emb}(\cdot)$ is a word embedding function. For MAD, we can obtain a new attribute set containing 5,801 \emph{single} attributes by treating the attributes in each composition as a whole. After the changes are made at the input level, most of the rest can be run directly according to the original methods, except for some methods that require minor modifications to ensure that they can work properly. For example, the dimension of embedding space of AttOperator \cite{nagarajan2018attributes} is changed to 100 to fit the GPU memory, and node pairs $(a_1, a_2)$, $(a_1, o)$, $(a_1, (a_1,a_2,o))$, $(a_2, o)$, $(a_2, (a_1,a_2,o))$, $(o, (a_1,a_2,o))$ are connected for CGE \cite{naeem2021learning} if $(a_1,a_2,o)$ is in the composition label set. All baselines are undertaken in the generalized setup \cite{purushwalkam2019task-driven} and implemented with the public code: \textit{\url{https://github.com/ExplainableML/czsl}}.

\subsection{Implementation Details}
We use the ResNet-18 \cite{he2016deep} network pre-trained on the ImageNet \cite{deng2009imagenet:} dataset to extract the 512-dimensional initial visual features of input images. For a fair comparison, neither a fine-tuning operation (including CGE \cite{naeem2021learning}) nor data augmentation is applied to our method and the baseline methods. We use the GloVe \cite{pennington2014glove:} model to extract the 300-dimensional linguistic attribute and object features. The linguistic encoder (i.e., GCN) is composed of two convolutional layers with the output channel numbers 4096 $\rightarrow$ 4096. For all three datasets, the initial learning rate is 1e-4 and the model is trained with batch-size 512 for 200 epochs. The loss weight $\beta$ is 0.5 for MIT-States and MAD, and 2 for UT-Zappos50K. The loss weight $\gamma$ is 0.1 for MIT-States and MAD, and 0.2 for UT-Zappos50K. The temperature $\tau$ is 20 for MIT-States, 70 for MAD, and 100 for UT-Zappos50K. All of the experiments are performed on an Nvidia RTX 2080Ti GPU with the TensorFlow deep learning platform.

\subsection{Single-Attribute-Object Composition Recognition}
Tab.~\ref{tab:methods-comparison-single} shows the comparison results of our method with baseline methods. As we can see, our method consistently outperforms the state-of-the-art recognition accuracies on the MIT-States dataset, achieving 7.8\% (Test AUC@1), 4.5\% (Seen), 3.2\% (Unseen), and 5.2\% (HM) improvements over the second-best method CGE\cite{naeem2021learning}. Notably, both CGE\cite{naeem2021learning} and our method adopt the graph-based network architecture but our method achieves better results with smaller computational cost, which demonstrates the effectiveness of our method. Similar observations are confirmed on the UT-Zappos50K dataset. Though our method behaves worse on Val AUC, it achieves significant improvements over the second-best methods, especially on HM.

\begin{table}[!h]
	\centering
	\caption{Comparison results on AUC and newly proposed Hard \& Soft metrics on MAD (\%). We report the top-k (k=1,2,3) AUC on the validation/testing set, the best Seen, Unseen, HM, Hard and Soft metrics on the testing set. All methods use ResNet-18 as the backbone.}
	\label{tab:methods-comparison-multi}
	\setlength
	\tabcolsep{4pt}
	\large
	\renewcommand\arraystretch{1.3}{
		\resizebox{0.49\textwidth}{!}{
			\begin{tabular}{lccccccp{0.7cm}<{\centering}p{1.1cm}<{\centering}p{0.7cm}<{\centering}p{0.7cm}<{\centering}p{1.1cm}<{\centering}}
				\toprule
				\multirow{2}{*}{Methods}   
				&\multicolumn{3}{c}{Val AUC} & \multicolumn{3}{c}{Test AUC} 
				&\multirow{2}{*}{Seen}  &\multirow{2}{*}{Unseen}  &\multirow{2}{*}{HM}  &\multirow{2}{*}{Hard}  &\multirow{2}{*}{Soft}  \\
				\cmidrule(lr){2-4}  \cmidrule(lr){5-7} 
				& 1 & 2 &3 & 1 & 2 &3   \\ 
				\cmidrule(lr){1-1} \cmidrule(lr){2-4}  \cmidrule(lr){5-7}  \cmidrule(lr){8-12} 
				AttOperator \cite{nagarajan2018attributes} &0.5 &1.2 &2.1 &0.3 &0.9 &1.6  &7.6 &5.8  &4.0  &16.4 &32.2 \\
				LabelEmbed+ \cite{nagarajan2018attributes} &2.8 &6.9 &11.4 &2.6 &6.7 &11.4  &21.9 &17.4  &12.5  &29.6 &52.0 \\
				GenModel \cite{nan2019recognizing} &5.1 &13.5 &21.7 &3.7 &11.3 &19.3  &18.2 &33.8  &12.3  &39.1 &68.0 \\
				TMN \cite{purushwalkam2019task-driven} &6.7 &17.6 &28.4 &5.4 &14.4 &23.0  &45.8 &18.4  &16.6  &43.9 &64.8 \\
				SymNet \cite{li2020symmetry} &18.0 &35.5 &48.2 &16.6 &33.3 &45.4  &74.4 &33.7  &29.7  &55.1 &73.5 \\
				CompCos \cite{mancini2021open} &18.0 &33.8 &46.6 &12.1 &26.5 &37.5 &\textbf{88.4} &19.7 &22.7 &\textbf{60.8} &\textbf{77.6}  \\
				CGE \cite{naeem2021learning} &19.6 &40.6 &55.6 &18.6 &39.1 &53.1 &56.7 &\textbf{43.6} &33.3 &53.6 &74.7  \\
				\cmidrule(lr){1-1} \cmidrule(lr){2-4}  \cmidrule(lr){5-7} \cmidrule(lr){8-12} 
				Ours &\textbf{24.2}  &\textbf{48.1} &\textbf{63.3} &\textbf{21.4} &\textbf{44.8} &\textbf{59.8} &68.5 &43.0  &\textbf{35.4}  &58.8  &77.2 \\ 
				\bottomrule
	\end{tabular}}}
\end{table}

\subsection{Multi-Attribute-Object Composition Recognition}
On the MAD dataset, we report the results on single-attribute metrics as well as the proposed multi-attribute metrics, with the former focusing on visibility and the latter emphasizing compositionality. As shown in Tab. \ref{tab:methods-comparison-multi}, our method outperforms all of the baseline methods on Val/Test AUC by large margins, achieving 23.4\% (Val AUC@1) and 15.1\% (Test AUC@1) improvements over the second-best method CGE \cite{naeem2021learning}. Although CompCos\cite{mancini2021open} achieves extremely high accuracy on Seen, it performs terribly on Unseen, which results in low HM scores. In contrast, our method achieves very competitive results on the single-attribute metrics. Besides, we report the comparison results on conventional dataset splits \cite{nagarajan2018attributes} in Suppl Section A.

As explained in Section \ref{sec_baselines_metrics}, the single-attribute metrics are insufficient to reflect the compositional nature in the multi-attribute case, which means that a method performing well on single-attribute metrics may not perform consistently well on multi-attribute metrics. As can be seen, SymNet\cite{li2020symmetry} achieves better results on single-attribute metrics than CompCos\cite{mancini2021open}, but it behaves worse on Hard and Soft metrics. In contrast, our method achieves excellent results on both single-/multi-attribute metrics, both from the perspective of seen/unseen domain differences and from the compositional characteristics of attributes and objects. Although CompCos\cite{mancini2021open} slightly outperforms our method on multi-attribute metrics, it cannot perform well on both single-/multi-attribute metrics. Further experiments about the calibration factor $\eta$ in multi-attribute metrics are conducted in Section \ref{sec_calibration_factor}, which shows that our method achieves superior results across a wide range of $\eta$.

\begin{table*}[b]
	\centering
	\caption{ Effects of different graph construction methods on the MIT-States, UT-Zappos50K and MAD datasets (\%).
}
	\label{tab:ablation-graph}
	\setlength
	\tabcolsep{7pt}
	\scriptsize
	\renewcommand\arraystretch{1.1}{
		\resizebox{0.96\textwidth}{!}{
			\begin{tabular}{p{.7cm}p{1.2cm}<{\centering}ccccccccccc}
				\toprule
				\multirow{2}{*}{Graphs} &\multirow{2}{*}{\tabincell{c}{Adj. matrix \\ trainable}}  &\multicolumn{3}{c}{MIT-States} & \multicolumn{3}{c}{UT-Zappos50K} &\multicolumn{5}{c}{MAD}  \\
				\cmidrule(lr){3-5} \cmidrule(lr){6-8}  \cmidrule(lr){9-13}
				&  &Val@1  &Test@1 &HM &Val@1 &Test@1 &HM &Val@1 &Test@1 &HM &Hard &Soft\\
				\cmidrule(lr){1-2} \cmidrule(lr){3-5} \cmidrule(lr){6-8}  \cmidrule(lr){9-13}
				$\mathcal{G_L}$ &$\times$ &4.9 &3.8 &14.6 &31.5 &26.4 &40.5 &14.7 &14.6 &29.4 &50.6 &71.8\\
				$\mathcal{G_E}$ &$\times$ &6.7 &5.3 &17.9 &23.6 &25.1 &42.1 &23.6 &20.8 &35.0 &58.0 &76.8\\
				$\mathcal{G_R}$ &$\checkmark$ &6.6 &5.2 &17.4 &29.4 &28.1 &44.3 &23.7 &21.1 &35.0 &58.3 &76.6\\
				$\mathcal{G_{SR}}$ &$\checkmark$ &6.8  &5.5 &18.1 &27.7 &33.5 &49.4 &24.2 &21.4 &35.4 &58.8 &77.2\\
				\midrule 
				\multicolumn{2}{c}{Non-graph} &6.4 &5.2 &17.0 &22.4 &24.4 &41.3 &22.5 &20.3 &34.5 &57.2 &76.1 \\
				\bottomrule
	\end{tabular}}}
\end{table*}

\subsection{Modeling Attribute and Object Relations}\label{sec_graph_relation}
In this experiment, we aim to explore a reasonable graph structure to model the relations between the attributes and objects. To compare with the four graph-based models (as described in Section \ref{sec_graph}), we introduce a non-graph model by setting the adjacency matrix of the proposed model to an identity matrix (no message passing between nodes). As shown in Tab.~\ref{tab:ablation-graph}, the models with different graphs basically achieve satisfactory accuracies and the graph models with a learnable adjacency matrix usually perform better than the graph models with a fixed adjacency matrix. After constraining the random graph $\mathcal{G_R}$ with the L1 norm, we observe that the sparse random graph $\mathcal{G_{SR}}$ achieves slightly better performance than that of the vanilla random graph, proving that the sparse connection between the attribute and object nodes is beneficial to learning good node representations. In addition, we observe that the link graph $\mathcal{G_L}$ achieves much worse performance among all of the graph methods, which is due to the introduction of noisy connections in the graph. Especially, for the MIT-States and MAD datasets with many attributes, objects, and compositions, it is easy to cause over-connections between nodes in the link graph $\mathcal{G_L}$. Importantly, compared with the other graph-based models, the recognition performance of the non-graph model is usually reduced. Summarizing the above, a sparsely connected attribute-object semantic association graph contributes to modeling meaningful relations and learning good node representations, but a dense graph could introduce noisy connections and harm performance.

\subsection{Ablation Studies for Loss Functions}
\begin{table*}[!t]
	\centering
	\caption{Effects of different loss function compositions (\%). We also consider replacing the proposed loss with other possible objectives.}
	\label{tab:loss-functions}
	\setlength
	\tabcolsep{7pt}
	\scriptsize
	\renewcommand\arraystretch{1.1}{
		\resizebox{0.96\textwidth}{!}{
			\begin{tabular}{lccccccccccc}
				\toprule
				\multirow{2}{*}{Methods} &\multicolumn{3}{c}{MIT-States} &\multicolumn{3}{c}{UT-Zappos50K} &\multicolumn{5}{c}{MAD}  \\
				\cmidrule(lr){2-4} \cmidrule(lr){5-7}  \cmidrule(lr){8-12}
				&Val@1  &Test@1 &HM &Val@1 &Test@1 &HM &Val@1 &Test@1 &HM &Hard &Soft\\
				\cmidrule(lr){1-1} \cmidrule(lr){2-4} \cmidrule(lr){5-7}  \cmidrule(lr){8-12}
				full model &6.8  &5.5 &18.1 &27.7 &33.5 &49.4 &24.2 &21.4 &35.4 &58.8 &77.2\\
				w/o $\mathcal{L}_{de}$ &6.3  &5.2 &17.7 &27.4 &27.5 &43.6 &23.5 &20.7 &35.0 &58.1 &76.7\\
				w/o $\mathcal{L}_{bal}$ &6.5 &5.2 &18.0 &14.0 &19.5 &36.2 &23.5 &20.5 &34.5 &58.3 &76.8 \\
				w/o $\mathcal{L}_{de}$ \& $\mathcal{L}_{bal}$ &6.2 &5.1 &17.5 &13.5 &19.3  &35.4 &23.2 &20.3 &34.2 &57.7 &76.3 \\
				w/o composer module &5.8  &4.4 &16.2 &27.4 &27.4 &43.1 &16.8  &15.4 &30.7 &53.4  &74.0  \\
				\cmidrule(lr){1-1} \cmidrule(lr){2-4} \cmidrule(lr){5-7}  \cmidrule(lr){8-12}
				$\mathcal{L}_{fus}$ with $L_1$ loss &2.3  &1.6 &8.9 &16.5 &13.3 &28.0  &5.5 &4.8 &15.9 &38.8 &60.3  \\
				$\mathcal{L}_{fus}$ with $L_2$ loss &2.8  &2.1 &10.5 &14.7 &15.8 &30.8 &4.4 &3.9 &13.5 &38.3 &60.0 \\
				\cmidrule(lr){1-1} \cmidrule(lr){2-4} \cmidrule(lr){5-7}  \cmidrule(lr){8-12}
				$\mathcal{L}_{bal} \rightarrow \mathcal{L}_{bal}^{'}$ &5.7  &2.8 &12.3 &14.3 &11.0 &25.5 &23.2 &20.0 &34.2 &58.2 &76.6 \\
				$\mathcal{L}_{bal} \rightarrow \mathcal{L}_{bal}^*$ \cite{xie2020region} &5.9  &4.8 &16.7 &14.9 &14.1 &28.4 &23.2 &20.7 &34.9 &58.4 &76.9 \\
				\cmidrule(lr){1-1} \cmidrule(lr){2-4} \cmidrule(lr){5-7}  \cmidrule(lr){8-12}
				$\mathcal{L}_{de}$ with $L_1$ loss &6.2  &4.9 &16.9 &25.8 &28.9 &45.2 &23.7 &20.6 &34.9 &58.3 &76.9 \\
				$\mathcal{L}_{de}$ with huber loss &6.6  &5.3 &17.9 &26.4 &30.0 &47.4 &23.6 &21.1 &35.4 &58.5 &77.0 \\
				$\mathcal{L}_{de}$ with MSLE loss &6.6  &5.2 &17.7 &26.6 &30.5 &46.1 &24.0 &21.3 &35.6 &58.5 &77.0 \\
				\bottomrule	
	\end{tabular}}}
\end{table*}

\begin{table*}[!t]
	\centering
	\caption{ Comparison results of the proposed model with different network architectures of feature extractors (\%). We compare the performance of different visual feature extractors and linguistic feature extractors.}
	\label{tab:feature-extractor}
	\setlength
	\tabcolsep{7pt}
	\scriptsize
	\renewcommand\arraystretch{1.1}{
		\resizebox{0.96\textwidth}{!}{
			\begin{tabular}{llccccccccccc}
				\toprule
				\multirow{2}{*}{\tabincell{c}{Feature \\ extractors}} & \multirow{2}{*}{\tabincell{c}{Network \\ architectures} } &\multicolumn{3}{c}{MIT-States} & \multicolumn{3}{c}{UT-Zappos50K} &\multicolumn{5}{c}{MAD}  \\
				\cmidrule(lr){3-5} \cmidrule(lr){6-8}  \cmidrule(lr){9-13}
				& &Val@1  &Test@1 &HM &Val@1 &Test@1 &HM &Val@1 &Test@1 &HM &Hard &Soft \\
				\cmidrule(lr){1-2} \cmidrule(lr){3-5} \cmidrule(lr){6-8}  \cmidrule(lr){9-13}
				\multirow{6}{*}{\tabincell{c}{Visual \\ feature}} 
				&VGG-16 &4.8  &3.5 &14.1 &24.7 &25.3 &41.5 &20.0 &19.2 &34.2 &55.8 &75.9 \\
				&VGG-19 &4.9 &3.9 &15.0 &25.4 &26.8 &42.7 &20.6 &19.9 &34.7 &56.4 &76.5 \\
				&ResNet-18 &6.8  &5.5 &18.1 &27.7 &33.5 &49.4 &24.2 &21.4 &35.4 &58.8 &77.2\\
				&ResNet-50 &8.5 &6.7 &20.2 &29.9 &34.6 &50.7 &26.3 &23.9 &38.3 &60.2 &79.8 \\
				&ResNet-101 &8.8 &6.8 &20.4 &30.8 &35.8 &51.1 &27.8 &25.6 &40.0 &61.3 &80.9 \\
				&GoogleNet &5.3 &4.4 &16.0 &22.7 &24.3 &41.6 &20.7 &18.5 &33.1 &55.9 &74.7 \\
				\cmidrule(lr){1-2} \cmidrule(lr){3-5} \cmidrule(lr){6-8}  \cmidrule(lr){9-13}			
				\multirow{4}{*}{\tabincell{c}{Linguistic \\ feature}}  
				&Glove   &6.8  &5.5 &18.1 &27.7 &33.5 &49.4 &24.2 &21.4 &35.4 &58.8 &77.2 \\
				&Word2Vector &6.5 &5.3 &17.6 &26.6 &29.4 &46.3 &22.2 &20.0 &34.4 &57.2 &76.3  \\
				&fastText &6.7 &5.4 &17.9 &30.5 &30.3 &45.9 &22.5 &20.5 &34.8 &57.6 &76.6  \\
				&One-hot &6.4 &5.0 &17.3 &27.4 &28.5 &45.3 &21.4 &18.7 &33.3 &55.9 &75.6 \\
				\bottomrule
	\end{tabular}}}
\end{table*}

In this experiment, we evaluate the effects of different loss function compositions, and the results are shown in Tab.~\ref{tab:loss-functions}. When the decoding loss $\mathcal{L}_{de}$ is removed from the full model, the recognition performance drops by 0.7\% to 17.9\%, which demonstrates that the decoding loss is beneficial to learn intrinsic features. If we disenable the balance loss $\mathcal{L}_{bal}$, we can see obvious degradations on the UT-Zappos50K dataset. If we remove the above two losses at the same time, the model shows the worst performance. Lastly, we validate the effectiveness of the attribute-object composer module. The experimental results show that a simple attention-based fusion scheme is capable of learning a holistic composition representation. It is also observed that the full model (with all losses and the composer module) obtains the highest performance, which further confirms the effectiveness of the proposed components. For more ablation studies about loss functions please refer to Suppl Section B.

Besides, we consider replacing the proposed losses with other possible loss functions. When the fusion loss $\mathcal{L}_{fus}$ is replaced with the L1 or L2 form, the model shows obvious degradations on all datasets. The main reason is that the contrastive form based fusion loss can maximize the similarity between the anchor image and the correct composition while minimizing the similarities with other incorrect compositions. Then, we replace the balance loss $\mathcal{L}_{bal}$ with $\mathcal{L}_{bal}^*$, which is proposed in \cite{xie2020region}. Moreover, we propose another formulation of balance loss: $\mathcal{L}_{bal}^{'}=- \frac 1 b \sum_{i=1}^b \max \limits _{k \in \mathcal{Y}^u} {\rm sim}(x_i, z_k)$, which aims to maximize the similarity between the anchor image and the most similar composition from the unseen compositions. As can be seen, neither $\mathcal{L}_{bal}^*$ nor $\mathcal{L}_{bal}^*$ can behave better than the naive $\mathcal{L}_{bal}$. The reason is that $\mathcal{L}_{bal}$ raises the model's awareness of all unseen compositions during the training, even for the least similar composition. Lastly, we validate the effectiveness of the visual decoder module. No matter which form of decoding loss is used, the model achieves satisfactory results.

\subsection{Image retrieval}
To qualitatively evaluate the attribute-object interdependent relations that our model has learned, we report the image retrieval results from a given attribute-object composition query. For each dataset, we use the trained model to retrieve the top-5 nearest images to the given query from the testing set. In practice, a query, which arises from various compositions, is unknown in advance. To better evaluate image retrieval results under different situations, the queries are categorized into three types. 1) Te: queries are chosen from the testing label set. 2) Tr: queries are selected from the training label set. 3) New: queries are synthesized manually by arbitrarily choosing attributes and objects and composing them to ensure that the manipulated compositions are neither in the training label set nor in the testing label set. Compared with the first two cases, there is no corresponding image in the testing set, and the query has not been seen. Therefore, it is very challenging for a model to retrieve reasonable images. Meanwhile, this experiment can test whether a model has essentially learned the concepts.

\begin{figure*}[!t]
	\centering
	\includegraphics[width=1\textwidth]{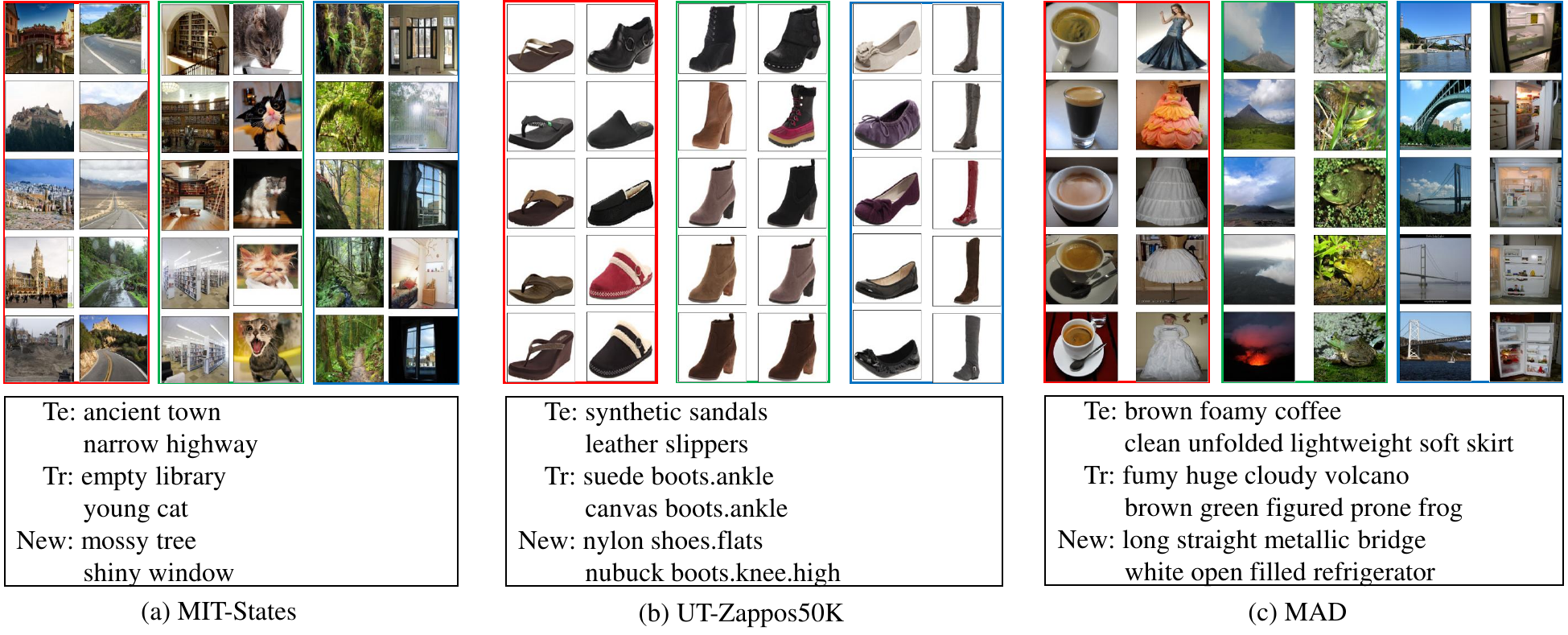}
	\caption{Image retrieval results. We show the image retrieval results given three different types of queries, i.e., "Te" (in red box), "Tr" (in green box) and "New" (in blue box). (a)-(c) The top-5 nearest images to the given query from the testing set are shown for the MIT-States, UT-Zappos50K and MAD datasets, respectively. The text below shows the corresponding queries, in order.
	}
	\label{fig_imgretrieval}
\end{figure*}

Image retrieval results are shown in Figure.~\ref{fig_imgretrieval}, where the first two columns represent the "te" queries, the middle two columns are "tr" queries, and the last two columns depict "new" queries for each dataset. All of the queries are listed in the box below the images, in order. Our proposed model is capable of retrieving a certain number of quite accurate images on the MIT-States (Figure.~\ref{fig_imgretrieval}a) and UT-Zappos50K (Figure.~\ref{fig_imgretrieval}b) datasets. The task on the MAD dataset is more difficult than that on the other two datasets because each query is composed of multiple attributes, which requires a deeper understanding of the attribute and object concepts and the relations between them. Admittedly, the results on this complex dataset (Figure.~\ref{fig_imgretrieval}c) suggest that our model has learned the inner concepts and interdependent relations between the attributes and objects.

\begin{figure*}[!t]
	\centering
	\includegraphics[width=\textwidth]{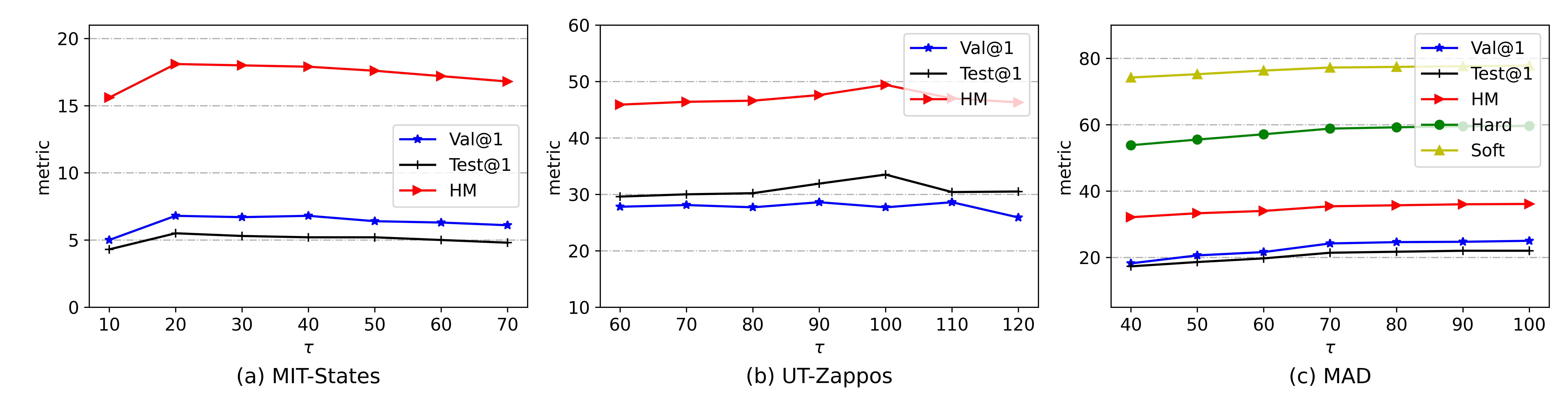}
	\caption{Impact of hyper-parameter $\tau$ on model performance.
	}
	\label{fig_para_tau}
\end{figure*}

\subsection{Visual and Linguistic Feature Extraction}
In Tab.~\ref{tab:feature-extractor}, we report performance corresponding to different types of image feature extractors and word embedding models. Firstly, we test three kinds of image feature extractors, VGG \cite{simonyan2014very}, ResNet \cite{he2016deep} and GoogleNet \cite{szegedy2015going}. As can be seen, visual features significantly affect recognition performance. Due to the residual connection \cite{he2016deep}, ResNet models usually behave better than VGG and GoogleNet, and they achieve higher performance with deeper architectures. Then, we study three widely used word embeddings including Glove \cite{pennington2014glove:}, Word2Vector \cite{mikolov2013efficient} and fastText \cite{joulin2017fasttext.zip:} to investigate the effects of different word embeddings. As shown in Tab.~\ref{tab:feature-extractor}, the Glove method achieves competitive performance on all three datasets. In addition to the word embedding models, we also use the one-hot encoding of labels as the inputs of the linguistic pathway. As a simple method, the one-hot encoding scheme behaves worse than those methods that employ word embeddings. The main reason is that the one-hot encoding does not consider the semantic relations between different composition labels.

\begin{table*}[!t]
	\centering
	\caption{ Effects of graph convolution depth (\%).}
	\label{tab:layer-number}
	\setlength
	\tabcolsep{7pt}
	\scriptsize
	\renewcommand\arraystretch{1}{
		\resizebox{0.96\textwidth}{!}{
			\begin{tabular}{cccccccccccc}
				\toprule
				\multirow{2}{*}{Layers} &\multicolumn{3}{c}{MIT-States} & \multicolumn{3}{c}{UT-Zappos50K} &\multicolumn{5}{c}{MAD}  \\
				\cmidrule(lr){2-4} \cmidrule(lr){5-7}  \cmidrule(lr){8-12}
				&Val@1  &Test@1 &HM &Val@1 &Test@1 &HM &Val@1 &Test@1 &HM &Hard &Soft \\
				\cmidrule(lr){1-1} \cmidrule(lr){2-4} \cmidrule(lr){5-7}  \cmidrule(lr){8-12}
				1 &6.4 &5.0 &17.1 &25.5 &29.7 &46.5 &19.6 &18.5 &33.2 &54.8 &74.8  \\
				2 &6.8  &5.5 &18.1 &27.7 &33.5 &49.4 &24.2 &21.4 &35.4 &58.8 &77.2\\
				3 &6.1 &4.8 &16.6 &25.8 &30.1 &47.3 &25.1 &21.7 &36.0 &59.6 &77.8  \\
				4 &5.8 &4.6 &16.5 &22.6 &28.6 &45.5 &24.3 &21.4 &35.5 &59.1 &77.3   \\
				\bottomrule
	\end{tabular}}}
\end{table*}

\subsection{Hyper-parameter Sensitivity Analysis}

\subsubsection{Impact of temperature $\tau$}
We explore the impact of hyper-parameter $\tau$ in Eq.~\ref{eq_fus} on the performance of the model in this experiment. As an important parameter in contrastive loss, $\tau$ controls feature distribution on the unit sphere. As shown in Figure.~\ref{fig_para_tau}, the model is robust to $\tau$ across a wide range of changes. As can be seen, good results can be obtained when $\tau$ is 20 for MIT-States, 100 for UT-Zappos50K, and 70 for MAD.

\subsubsection{Depth of CGN}
Besides, the robustness of the model to the number of layers is examined in Tab.~\ref{tab:layer-number}. The 3-layer model has the output dimensions as 4096 $\rightarrow$ 4096 $\rightarrow$ 4096 and the 4-layer model has the output dimensions as 4096 $\rightarrow$ 4096 $\rightarrow$ 4096 $\rightarrow$ 4096. As can be seen, the recognition performance is robust to the depth of the model, even for the shallow 1-layer model (output dimension being 4096). The recognition performance raises as the depth of model increases from one layer to two layers. Because a deeper model involves more convolution operations over the graph, which are essential to aggregate information of distant nodes. Nevertheless, the performance reaches saturation and drops slightly when the depth of a model increases to a certain point, possibly due to the over smoothing \cite{li2018deeper,chen2020simple} problem of GCN.

\begin{figure}[!t]
	\centering
	\includegraphics[width=0.5\textwidth]{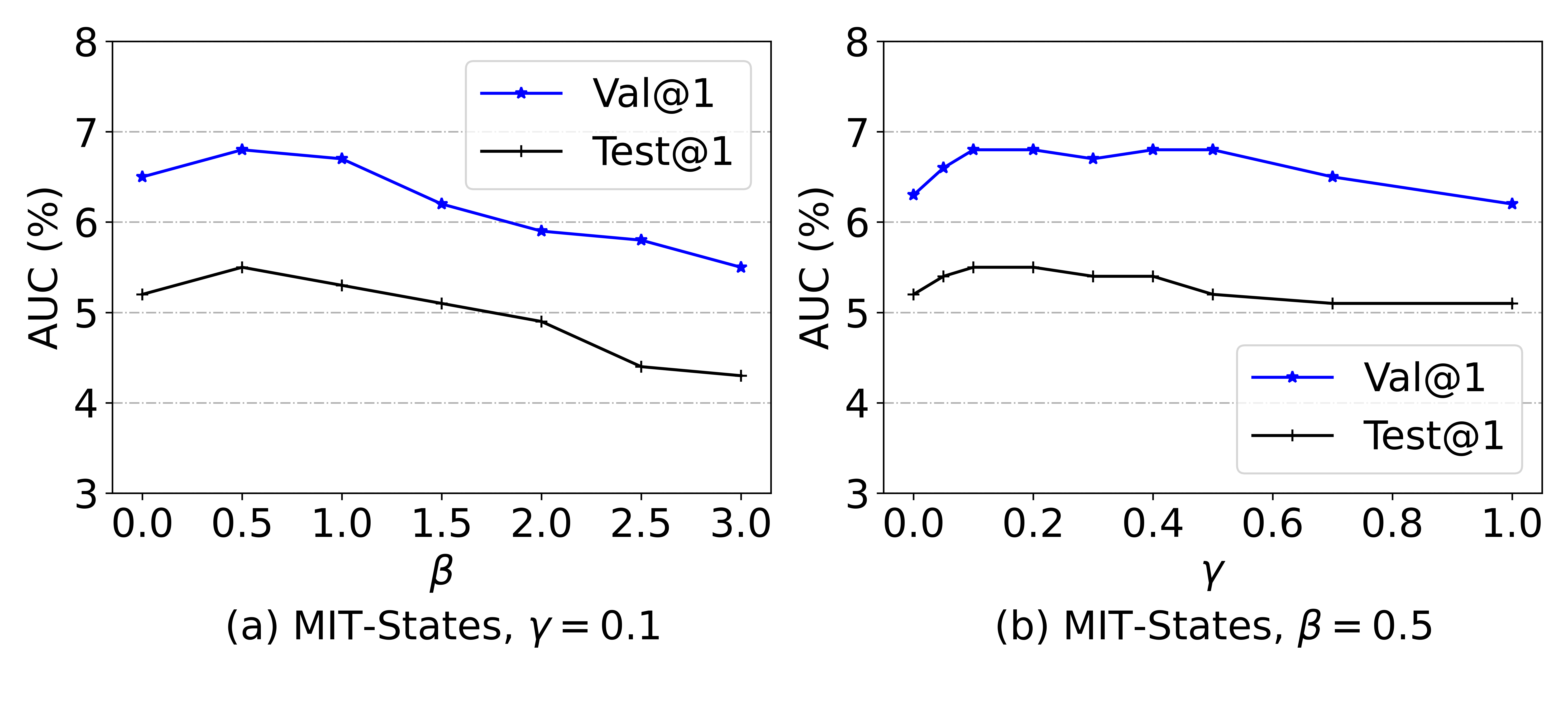}
	\caption{Effects of loss weights $\beta$ and $\gamma$ on model performance. 
		Note that $\beta=0.5, \gamma=0.1$ is our default hyper-parameter setting on the MIT-States dataset. 
	}
	\label{fig_loss_weights}
\end{figure}

\subsubsection{Effects of loss weights $\beta$ and $\gamma$}
The influences of loss weights $\beta$ and $\gamma$ in Eq.~\ref{eq_loss} are shown in Figure.~\ref{fig_loss_weights}. On the MIT-States dataset, AUC increases slightly as $\beta$ gradually increases from 0 to 0.5 when $\gamma$ is fixed to the default value of 0.1. However, as the balance loss gradually dominates the overall loss (with $\beta$ increasing), performance decreases rapidly. In addition, AUC increases as $\gamma$ gradually increases to 0.1 when $\beta$ is fixed to the default value of 0.5. Then, the performance reaches saturation and drops slightly as $\gamma$ continues to increase.

\begin{figure}[!t]
	\centering
	\includegraphics[width=0.46\textwidth, height=0.72\textwidth]{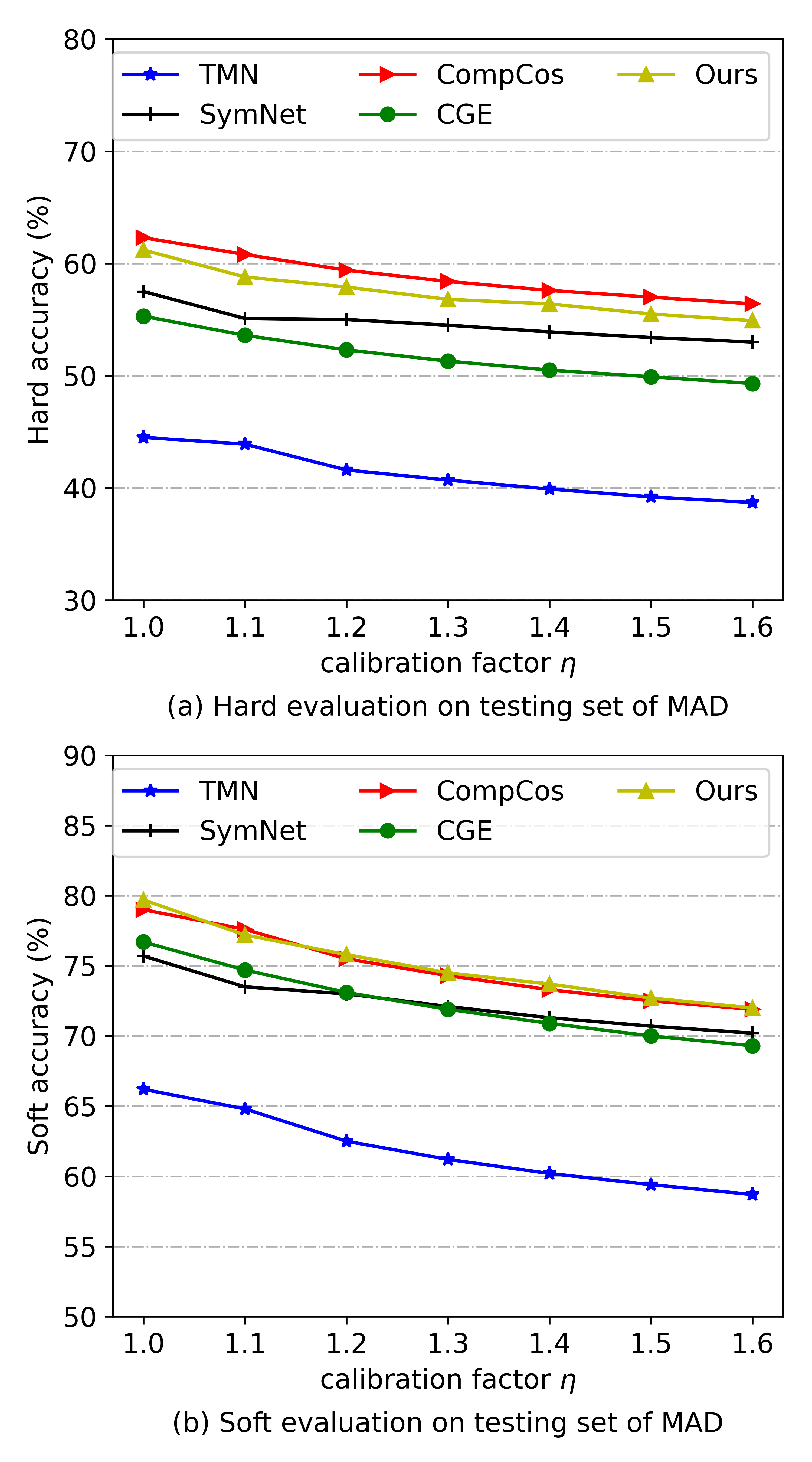}
	\caption{Effects of calibration factor $\eta$.
	}
	\label{fig_account_factor}
\end{figure}

\subsubsection{Effects of calibration factor $\eta$} \label{sec_calibration_factor}
Finally, we exhaustively compare the impact of the calibration factor $\eta$ on different methods on the MAD dataset. As a calibration term, $\eta$ penalizes predictions whose length exceeds ground-truth. Generally, a heavier penalty is imposed when $\eta$ increases. We plot the Hard/Soft accuracy curves for different $\eta$ in Figure.~\ref{fig_account_factor}. As can be seen, our method achieves very competitive results, and almost all methods show similar downward trends as $\eta$ increases.

%

\section{Conclusion}
This paper studies a problem called unseen attribute-object composition recognition. In view of the fact that images belonging to similar compositions usually lead to mis-classifications, we utilize the contrastive form based fusion loss to learn discriminative visual features. Meanwhile, the attributes and objects are interdependent, and the relations between them are captured via a graph neural network. The proposed model presents competitive performance and shows good flexibility and extensibility to handle the more challenging multi-attribute-object recognition problem. The quantitative results show that the proposed method improves the state-of-the-art performance effectively. More qualitative results illustrate that our model can learn the concepts of the attributes and objects and capture the complex associations between them. 

However, current progress is still far from human-like intelligence. There are at least three issues that deserve to be further studied in future work. 1) The currently proposed model is performed in a generalized setting, i.e., we know in advance which single-/multi-attribute compositions are seen or unseen, and do not predict compositions in an open-world setting \cite{mancini2021open}, which may restrict the applicability of the model in practice. 2) Estimation of attribute uncertainty: the attribute annotation task is subjective, and it could be difficult to determine whether an object contains an attribute or not. Thus, a practical way to eliminate this uncertainty (i.e., label ambiguity \cite{gao2017deep,2018Joint}) is to quantify the degree to which an object has a certain attribute, i.e., the confidence in the attribute. 3) Multi-Attribute Multi-Object Composition (MAMOC) recognition: the real world usually involves multiple objects and each object has multiple attributes. Therefore, a good MAMOC recognition model will pave the way for high-level AI.


%


\ifCLASSOPTIONcompsoc
  \section*{Acknowledgments}
\else
  \section*{Acknowledgment}
\fi

This work is supported by the National Science Foundation of China (No. 62088102).

\ifCLASSOPTIONcaptionsoff
  \newpage
\fi



%
%
%

\bibliographystyle{IEEEtran}
\bibliography{bibfile}

%

\begin{IEEEbiography}[{\includegraphics[width=1in,height=1.25in,clip,keepaspectratio]{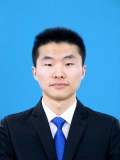}}]{Hui Chen} received the B.Sc. degree from Xi’an Jiaotong University, Xi’an, China, in 2019, where he is currently pursuing the Ph.D. degree. His current research interests include zero-shot learning, attribute recognition, graph neural networks and self-supervised learning methods.
\end{IEEEbiography}

\begin{IEEEbiography}[{\includegraphics[width=1in,height=1.25in,clip,keepaspectratio]{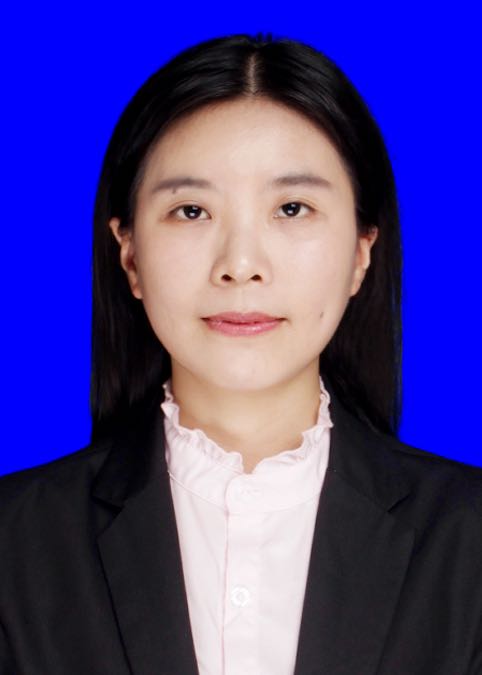}}]{Jingjing Jiang} received the B.S. degree in mechanical engineering from Xi'an Jiaotong University in 2017. She is currently a Ph.D. student with the Institute of Artificial Intelligence and Robotics of Xi'an Jiaotong University. Her research interests include computer vision and deep learning. 
\end{IEEEbiography}

\begin{IEEEbiography}[{\includegraphics[width=1in,height=1.25in,clip,keepaspectratio]{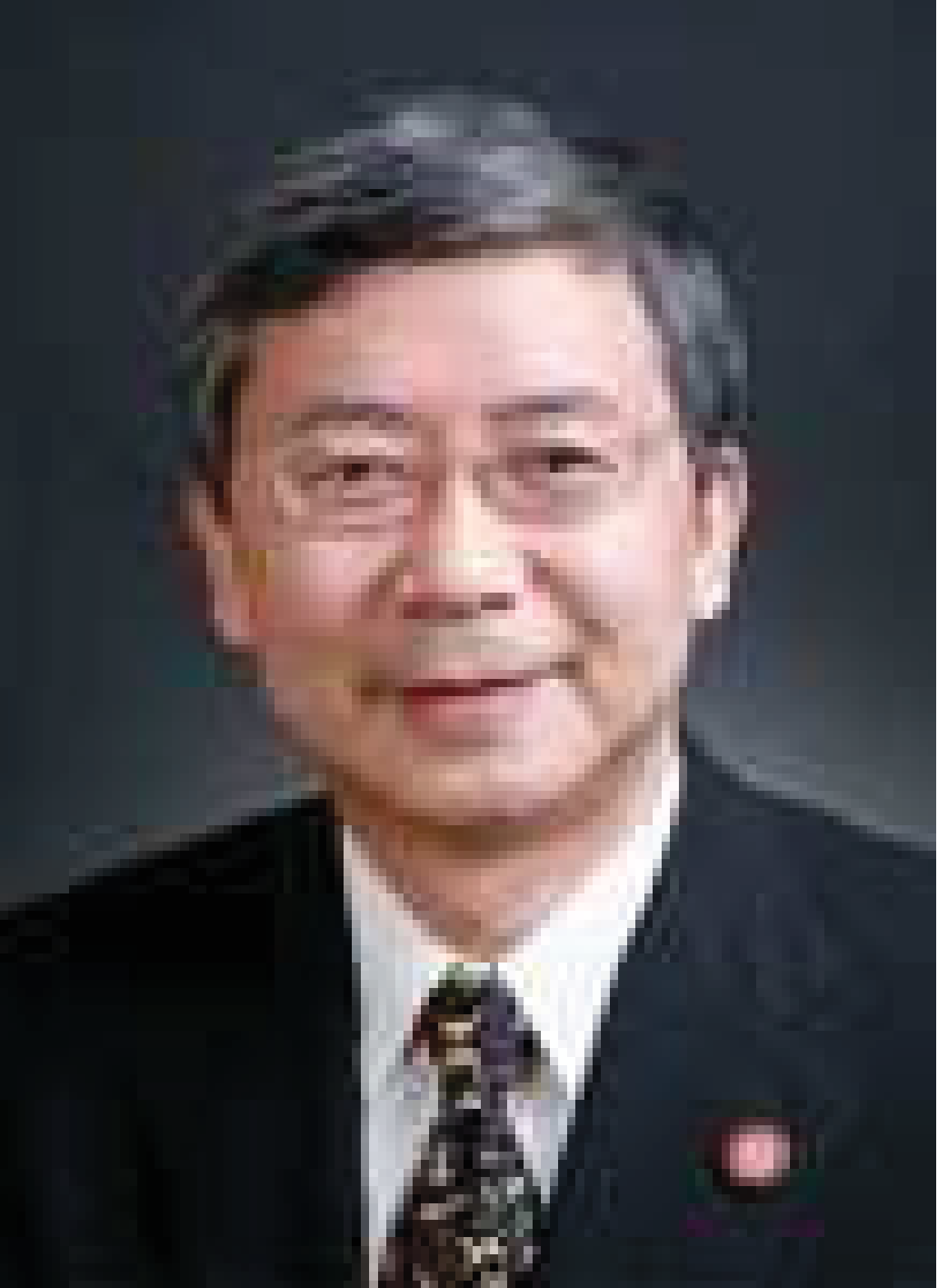}}]{Nanning Zheng}(Fellow, IEEE) received the bachelor’s degree from the Department of Electrical Engineering, Xi’an Jiaotong University, Xi’an, China, in 1975, the M.S. degree in information and control engineering from Xi’an Jiaotong University in 1981, and the Ph.D. degree in electrical engineering from Keio University, Yokohama, Japan, in 1985. He joined Xi’an Jiaotong University in 1975, where he is currently a Professor and the Director of the Institute of Artificial Intelligence and Robotics. His research interests include computer vision, pattern recognition, autonomous vehicle, and brain-inspired computing. He became a member of the Chinese Academy of Engineering in 1999. He is a Council Member of the International Association for Pattern Recognition.
\end{IEEEbiography}




\end{document}